\documentclass[10pt,twocolumn,letterpaper]{article}

\usepackage{paperlayout}

\usepackage{booktabs}
\usepackage{siunitx}
\usepackage{float}

\usepackage{cuted}

\newcommand{\ci}[2]{\,{\scriptsize[#1,\,#2]}}
\newcommand{\twoline}[2]{\begin{tabular}[t]{@{}l@{}}#1\\#2\end{tabular}}
\usepackage{xspace}
\newcommand{\methodname}{\textsc{SuperDyn}\xspace}
\newcommand{\projecturl}{https://superdyn.github.io/}

\definecolor{linkblue}{rgb}{0.21,0.49,0.74}
\usepackage[pagebackref,breaklinks,colorlinks,allcolors=linkblue]{hyperref}

\title{\methodname: Preserving Thin Structures in Superquadric
Decompositions of Deforming Objects}

\author{
  Nir Tevet$^{1}$ \qquad Yoni Kasten$^{1,2}$\\
  $^1$Technion\quad
  $^{2}$NVIDIA Research
}

\begin{document}
\maketitle

\begin{strip}
\centering
\includegraphics[width=\textwidth]{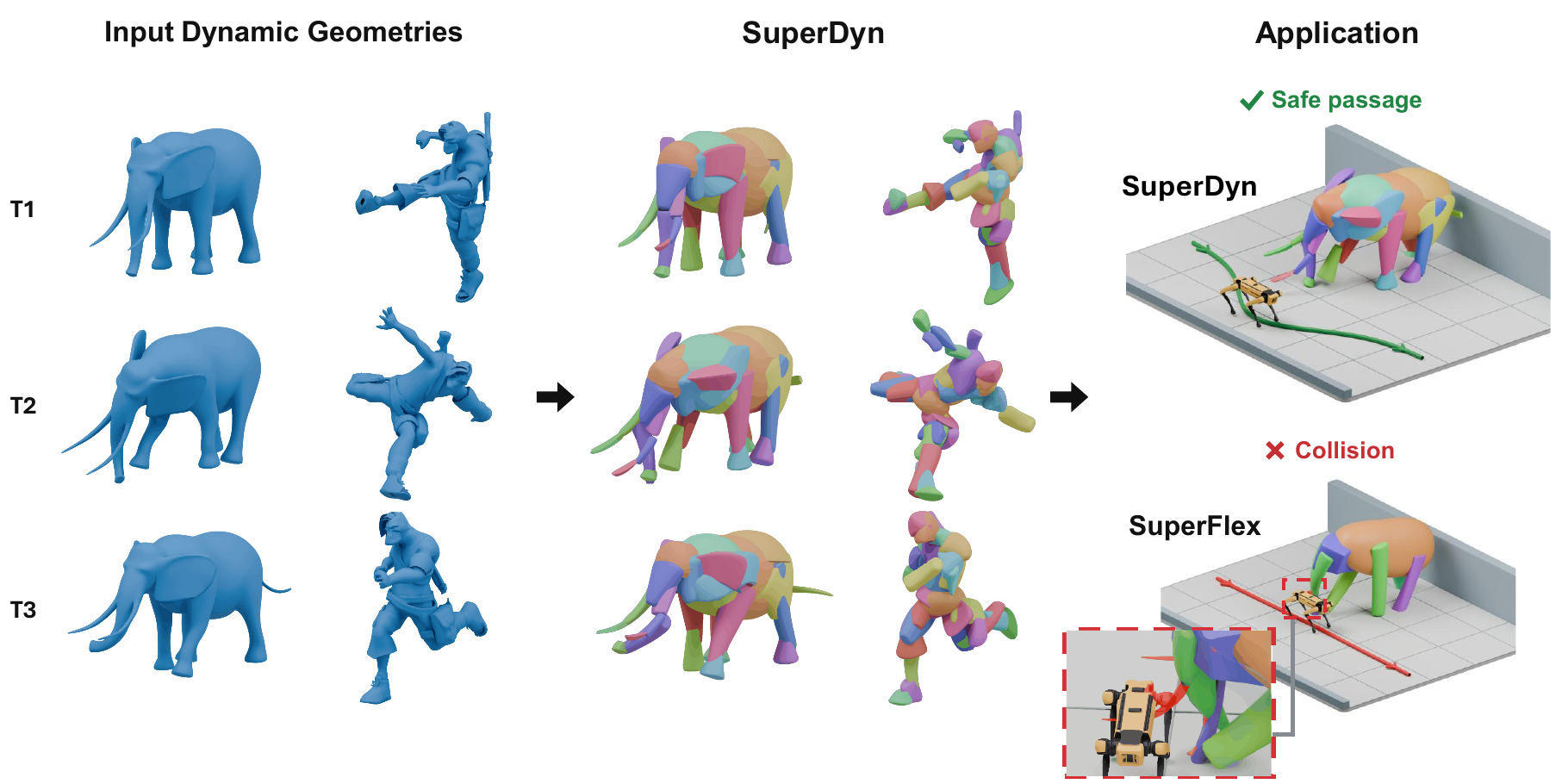}
\captionsetup{hypcap=false}
\captionof{figure}{\textbf{\methodname} decomposes a deforming mesh sequence into
superquadric primitives that keep their identities over time
(consistent colors across $T_1$--$T_3$) and preserve thin parts such as
tusks, tails, and limbs. When used as a collision proxy for motion
planning (robot rendered for illustration), \methodname yields a path that avoids the true geometry, whereas
SuperFlex misses the elephant's tusk (red, inset) and the planned path
collides with it.}
\label{fig:teaser}
\captionsetup{hypcap=true}
\end{strip}
\begin{abstract}
Simulation and robot navigation around deforming objects, such as animals and humans, require collision proxies that are compact, fast to build, and faithful to thin parts like limbs and tails, which often move fastest and sweep the most free space. Learned superquadric decompositions are a fast, compact alternative to approximate convex decomposition, but they tend to miss thin parts, a failure that volumetric IoU barely registers. We show that this failure matters: most collisions of invalid planned paths occur in thin regions, which cover only a fifth of the object surface. We present \methodname, a general approach that adapts feed-forward superquadric decomposition networks to deforming objects without adding network components. Interior-weighted occupancy supervision makes the network cover thin parts, and region-balanced input sampling keeps them covered under test-time refinement. Decomposing a single keyframe and moving each primitive rigidly through the sequence makes the decomposition temporally consistent. To measure thin-part coverage, we also introduce Thin-Structure Recall (TSR), which predicts planning success much better than IoU. On DeformingThings4D and Objaverse, \methodname roughly doubles TSR relative to SuperFlex. In motion planning around deforming obstacles, it reaches about 95\% success, compared with 83.7\% for the best primitive-based baseline and 87.1\% for rig-driven capsules, and is comparable to convex decomposition at 8--56$\times$ less storage. Applied to SuperDec, the same approach raises its planning success from 70.6\% to 93.8\%. We will release our code and a planning-based mini-benchmark for evaluating collision proxies of deforming obstacles. See our \href{\projecturl}{project page}.
\end{abstract}
\section{Introduction}
\label{sec:intro}

Physics simulation, games, and robot navigation increasingly involve deforming objects such as animals and humans. Each of these applications needs a collision proxy: a simplified volumetric representation that supports fast proximity queries. For deforming objects, a useful proxy must be compact enough to store for every frame, fast enough to build for many assets, and consistent over time. It must also cover thin parts such as limbs, tails, ears, and tusks. These parts move quickly and sweep through free space, so a proxy that misses them lets a planner route a path through occupied space (Fig.~\ref{fig:teaser}).

Existing proxies trade these requirements against each other. Approximate convex decomposition (ACD)~\cite{mamou2016volumetric,wei2022approximate,thul2018approximate} covers thin parts well, but it is slow to construct and produces large representations, since each frame or part is stored as a set of convex hulls. Rig-based capsules, the common practice in game engines, require a skeleton and skinning weights and still miss many thin parts. Learned superquadric decompositions~\cite{fedele2025superdec,tavernini2026superflex} predict a compact set of primitives in a single forward pass, but they decompose each frame independently and, as we show, systematically miss thin parts (Fig.~\ref{fig:overlay}).

This failure has received little attention, in part because standard reconstruction metrics barely register it: a missing tail or tusk changes volumetric IoU only slightly. We show that it nonetheless has a direct cost in planning. Across all methods we evaluate, 54--86\% of the collisions of invalid paths occur in thin regions, although these regions cover only about a fifth of the object surface. To measure thin-part coverage directly, we introduce Thin-Structure Recall (TSR), the fraction of an object's thin parts that a proxy covers, which we report alongside excess volume. Across methods, TSR is strongly correlated with planning success (Pearson $r=0.96$), whereas IoU is only moderately correlated with it ($r=0.49$; Fig.~\ref{fig:tsr-iou}).

We present \methodname, a general approach that adapts feed-forward superquadric decomposition networks to deforming objects without adding network components. Thin parts are missed because they receive few occupancy queries, few input points, and few primitives, and \methodname addresses each cause. Interior-weighted occupancy supervision draws more volumetric loss queries from inside the object, which makes the network cover thin parts. Region-balanced sampling partitions the object into regions around farthest-point anchors and draws an equal number of input points from each region, so that thin parts stay covered when the prediction is refined by test-time optimization (TTO). A larger primitive budget, without the sparsity term that favors few primitives, keeps primitives available for thin parts. To represent a whole sequence, the network decomposes a single keyframe, and each primitive is propagated through the sequence with a per-frame rigid transformation estimated from known point correspondences, so that primitive identities and shapes remain fixed over time. We apply \methodname to SuperFlex~\cite{tavernini2026superflex} and SuperDec~\cite{fedele2025superdec}. We provide a fast feed-forward variant and a slower variant that refines the keyframe with TTO.

On animated characters from DeformingThings4D~\cite{li20214dcomplete} and Objaverse~\cite{deitke2023objaverse}, \methodname roughly doubles TSR relative to SuperFlex, also at the same primitive budget, and remains clearly ahead when SuperFlex is dilated to the same excess volume. In motion planning around deforming obstacles, our feed-forward and TTO variants reach success rates of 94.8\% and 95.1\%, respectively, compared with at most 83.7\% for primitive-based baselines, and reduce collisions in thin regions by a factor of four. Both variants reach a success rate comparable to convex decomposition while requiring 8--56$\times$ less storage and constructing proxies more than 5$\times$ faster. Applied to SuperDec, \methodname raises planning success from 70.6\% to 93.8\%.

In summary, our contributions are:
\begin{itemize}
  \item We show that omitting thin parts, a failure that standard metrics overlook, causes most invalid planned paths around deforming obstacles, and introduce Thin-Structure Recall (TSR) to measure it.
 \item We introduce a planning-based mini-benchmark for evaluating collision proxies of deforming obstacles, which we will release together with our code.
  \item We present \methodname, a general approach that makes feed-forward superquadric decomposition networks cover thin parts and stay temporally consistent, without adding network components.
  \item Applied to SuperFlex and SuperDec, \methodname outperforms all primitive-based baselines in planning and reaches a success rate comparable to convex decomposition at a fraction of its storage and construction cost.
\end{itemize}
\section{Related Work}

\paragraph{Primitive-based decomposition.}
Superquadrics~\cite{barr1981superquadrics} describe a wide range of shapes with few parameters, and early work fitted them, with tapering and bending, to range data~\cite{solina1990recovery}. Optimization-based methods fit primitives directly to the input geometry, for example by probabilistic fitting to point clouds (EMS~\cite{liu2022robust}), from signed distance fields (Marching-Primitives~\cite{liu2023marching}), or to generated meshes~\cite{wang2025light}. Learning-based methods predict cuboids (CSA~\cite{yang2021unsupervised}) or superquadrics (SQ~\cite{paschalidou2019superquadrics}) without part annotations. SuperDec~\cite{fedele2025superdec} predicts superquadrics and point-to-primitive assignments in a feed-forward pass, and SuperFlex~\cite{tavernini2026superflex} adds tapering and bending. These learned models sample the input surface without regard to part size and are trained to favor few primitives, so thin parts are under-represented in both the input and the output. We address this imbalance for both SuperDec and SuperFlex.

\paragraph{Temporally consistent decomposition.}
Skinning methods keep part identities across poses through shared motion: SSDR~\cite{le2012smooth} fits rigid bone transformations and shared vertex weights to multiple poses of a mesh, and later work extracts a full skeletal rig from a mesh sequence~\cite{le2014robust}. D-Prism~\cite{yu2026d} reconstructs dynamic objects from monocular video with moving superquadrics that carry 3D Gaussians. Closer to our goal, Thul et al.~\cite{thul2018approximate} select mesh cuts across multiple poses and use the resulting partition to construct convex hulls for corresponding parts throughout the sequence. Animated sphere-meshes~\cite{thiery2016animated} approximate an animated surface with temporally coherent spheres for approximate collision detection. Both rely on per-sequence optimization, whereas we decompose a single keyframe in a feed-forward pass and move each primitive rigidly through the sequence.

\paragraph{Collision proxies.}
Approximate convex decomposition (ACD)~\cite{lien2007approximate} represents a mesh by the convex hulls of nearly convex parts, which supports fast collision queries in games and physics simulation. V-HACD~\cite{mamou2016volumetric} partitions a voxelization of the mesh, and CoACD~\cite{wei2022approximate} cuts the mesh guided by a collision-aware concavity metric. Sphere-trees~\cite{bradshaw2004adaptive} and sphere-meshes~\cite{thiery2013sphere} trade fidelity for fast proximity queries on static shapes, and medial-axis methods such as Coverage Axis~\cite{dou2022coverage} explicitly target thin structures. For rigged characters, game engines attach a capsule to each bone, which we include as a baseline (Sec.~\ref{sec:experiments}). A proxy can also close a passage that exists in the original geometry, which Navigation-Driven ACD~\cite{andrews2024navigation} prevents, or miss a thin part that a planned path crosses, which is the failure we target. Superquadric proxies have been used for grasping and motion planning~\cite{fedele2025superdec,vezzani2017grasping} and for reactive obstacle avoidance~\cite{ginesi2021dynamic,zhao2026sq}. We compare proxies of the same deforming obstacle under a shared planner.

\begin{figure*}
\centering
\includegraphics[width=\textwidth]{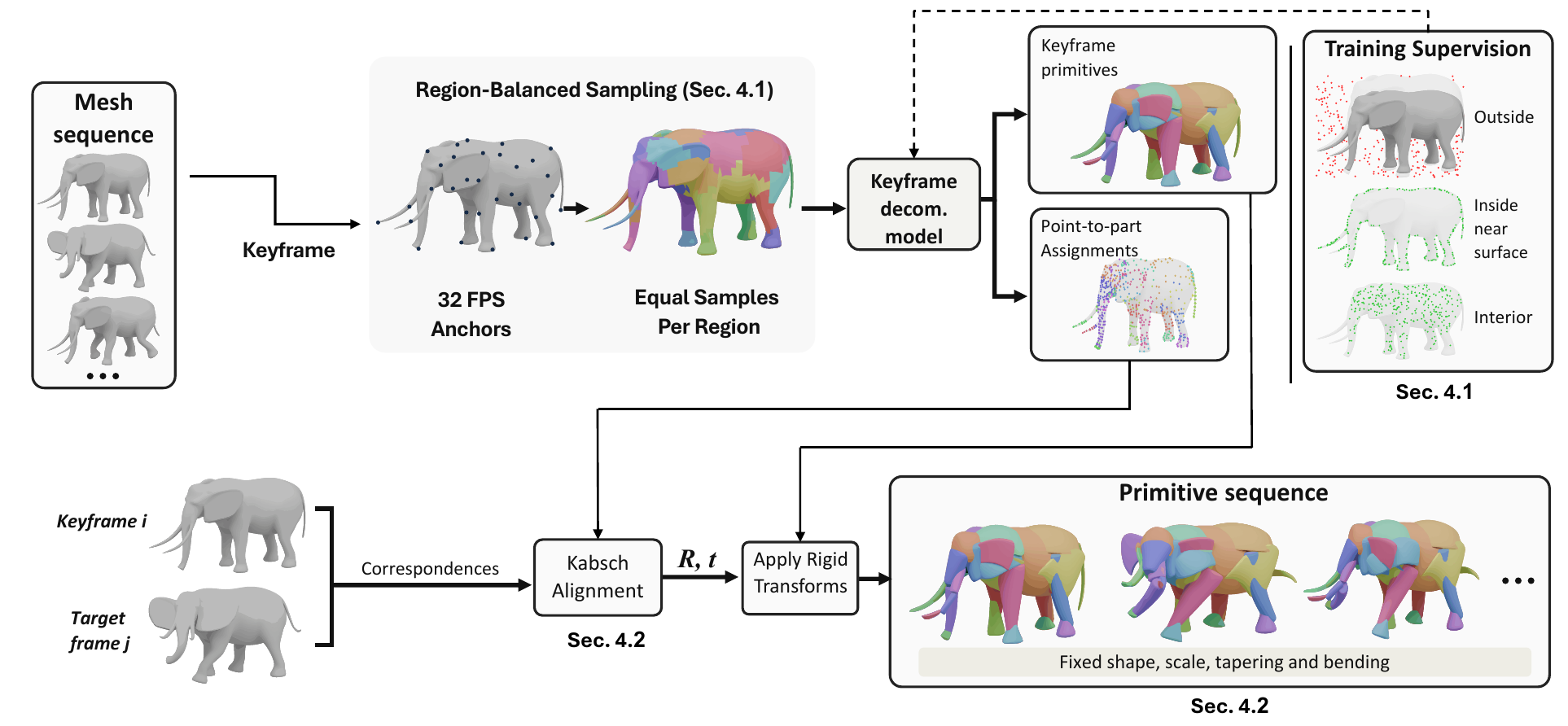}
\captionsetup{hypcap=false}
\captionof{figure}{\textbf{Pipeline overview.} \emph{Top:} we decompose a
keyframe of the mesh sequence (Sec.~\ref{sec:thin-parts}). Region-balanced
sampling selects 32 anchors by FPS and draws an equal number of input
points from each anchor's region (top left). The decomposition network
predicts the keyframe primitives and point-to-part assignments (top
center). During training, its occupancy loss uses queries outside the
object, inside it near the surface, and in its interior (top right).
\emph{Bottom:} for each primitive, Kabsch alignment of its assigned
points with their correspondences in a target frame gives a rigid
transform $(\mathbf{R},\mathbf{t})$ (bottom left), which moves the
primitive while its shape, scale, tapering, and bending stay fixed
(bottom right; Sec.~\ref{sec:key-frame-to-sequence}).}
\label{fig:our-method}
\captionsetup{hypcap=true}
\end{figure*}
\section{Preliminaries}
\label{sec:prelim}

\subsection{Superquadric Primitives}
\label{sec:prelim-sq}

A superquadric is a simple shape that can look like a box, a cylinder,
an ellipsoid, or something in between~\cite{barr1981superquadrics}. In
its local frame, it is the solid $F(\mathbf{x})\le 1$~\cite{barr1981superquadrics,liu2022robust}, where
\begin{equation}
\begin{aligned}
F(\mathbf{x}) ={}&
\left(
\left|\frac{x}{s_x}\right|^{2/\epsilon_2}
+
\left|\frac{y}{s_y}\right|^{2/\epsilon_2}
\right)^{\epsilon_2/\epsilon_1} \\
&+ \left|\frac{z}{s_z}\right|^{2/\epsilon_1}.
\end{aligned}
\end{equation}
The scales $\mathbf{s}$ set its size along each axis, and the exponents
$\boldsymbol{\epsilon}$ set how round it is: small values give boxy
shapes, and values near $1$ give round ones.

Many object parts are not straight or evenly thick: a tail narrows
toward its tip, and an elephant's trunk curves.
SuperFlex~\cite{tavernini2026superflex} therefore adds two
deformations~\cite{barr1984global}. \emph{Tapering}, with parameters $\boldsymbol{\tau}$,
makes a primitive thicker at one end than at the other, like a cone.
\emph{Bending}, with parameters $\boldsymbol{\beta}$, curves its axis
along a circular arc, like a curved tail. Both are applied in the
primitive's local coordinate system, and a rotation $\mathbf{R}$ and a translation
$\mathbf{t}$ then place it in the object. We refer
to~\cite{tavernini2026superflex} for the exact form of the
deformations. In total, primitive $k$ is described by
\begin{equation}
\boldsymbol{\theta}_k =
\bigl(
\mathbf{s}_k, \boldsymbol{\epsilon}_k,
\boldsymbol{\tau}_k, \boldsymbol{\beta}_k,
\mathbf{R}_k, \mathbf{t}_k, e_k
\bigr),
\end{equation}
where $e_k$ is an existence score that indicates whether the primitive
is used. SuperDec~\cite{fedele2025superdec} uses undeformed primitives,
which corresponds to $\boldsymbol{\tau}_k=\boldsymbol{\beta}_k=\mathbf{0}$.

\subsection{Feed-Forward Superquadric Decomposition}
\label{sec:prelim-ff}

SuperDec and SuperFlex decompose a shape in a single forward pass. From
points $\mathbf{X}\in\mathbb{R}^{N\times 3}$ sampled on the surface by
global farthest point sampling (FPS)~\cite{qi2017pointnet++}, a transformer predicts $K=16$
candidate primitives, one per part token, and a soft assignment matrix
$\mathbf{A}\in\mathbb{R}^{N\times K}$ that says how strongly each point
belongs to each primitive. The existence scores switch off unneeded
candidates.

SuperFlex is trained with
\begin{equation}
\begin{aligned}
\mathcal{L} ={}&
\lambda_{\mathrm{IoU}} \mathcal{L}_{\mathrm{IoU}}
+ \lambda_{\mathrm{SDF}} \mathcal{L}_{\mathrm{SDF}}
+ \lambda_{\mathrm{overlap}} \mathcal{L}_{\mathrm{overlap}} \\
&+ \lambda_{\mathrm{exist}} \mathcal{L}_{\mathrm{exist}}
+ \lambda_{\mathrm{SPS}} \mathcal{L}_{\mathrm{SPS}}.
\end{aligned}
\label{eq:superflex-loss}
\end{equation}
$\mathcal{L}_{\mathrm{IoU}}$ makes the primitives fill the object's
volume, using occupancy queries drawn uniformly from the unit cube
around the object. $\mathcal{L}_{\mathrm{SDF}}$ pulls the primitive
surfaces toward the input points, $\mathcal{L}_{\mathrm{overlap}}$
discourages overlapping primitives, $\mathcal{L}_{\mathrm{exist}}$
supervises the existence scores, and $\mathcal{L}_{\mathrm{SPS}}$
favors few primitives. A thin part fills only a tiny share of the
cube, so few queries land inside it, and $\mathcal{L}_{\mathrm{IoU}}$
gives little signal to cover it (Sec.~\ref{sec:thin-parts}). SuperDec
does not use occupancy queries. It is trained to bring the primitive
surfaces close to the input points, and it also supervises existence
and favors few primitives~\cite{fedele2025superdec}.

At test time, SuperFlex can refine the predicted primitive parameters of a single shape by test-time optimization (TTO) with the same loss while the network weights are fixed. On a sequence, both
models decompose each frame separately, so the result is not temporally
consistent. The same primitive may cover the tail in one frame and a leg
in the next, and the number and shapes of the primitives can change
between frames.

\section{Method}
\label{sec:method}
\methodname adapts any feed-forward network that maps a surface point cloud $\mathbf{X}$ to $K$ primitives $\boldsymbol{\theta}_k$ and point-to-part assignments $\mathbf{A}$ (Sec.~\ref{sec:prelim-ff}) and changes only its number of part tokens. In the top row of Fig.~\ref{fig:our-method}, region-balanced sampling selects the input points of a keyframe, and the network, trained with interior-weighted occupancy supervision, predicts its primitives and assignments (Sec.~\ref{sec:thin-parts}). In the bottom row, each primitive moves rigidly to every other frame by aligning its points with their correspondences, which yields a temporally consistent sequence (Sec.~\ref{sec:key-frame-to-sequence}).

\subsection{Covering Thin Parts}
\label{sec:thin-parts}
\label{sec:FPS-anchors}
\label{sec:training-objective}

Thin parts are easy to miss for three reasons: they receive few input
points, few occupancy queries, and few primitives. We address each with
one change.

\paragraph{Region-balanced input sampling.}
Global FPS spreads the input points evenly over the surface, so a thin
part, which has little surface area, receives few of them. We instead
divide the surface into $L$ regions and give each region the same number
of points (Fig.~\ref{fig:our-method}, top left). We select $L$ anchor
vertices of the keyframe mesh by FPS, which spreads them roughly evenly
over the surface, assign every vertex to its nearest anchor, and sample
$N/L$ vertices uniformly from each region. A region on a thin part has
much less surface than a region on the body, so it receives more points
per unit area. We use the same sampling during training and inference.
As our ablations show (Sec.~\ref{sec:ablation}), this mainly keeps thin
parts covered during TTO.

\paragraph{Interior-weighted occupancy supervision.}
The occupancy loss $\mathcal{L}_{\mathrm{IoU}}$ checks at query points
whether the primitives cover the object. SuperFlex draws these queries
uniformly from the unit cube, but a character fills only a small part of
the cube, and its thin parts an even smaller one, so few queries land
inside them. We instead draw $75\%$ of the queries outside the object,
$12.5\%$ inside it near the surface, and $12.5\%$ throughout its
interior (Fig.~\ref{fig:our-method}, top right). A fixed
share of the queries now lands inside the object, including its thin
parts, which pushes the network to cover them. For a network trained
without an occupancy loss, such as SuperDec, we fine-tune with the
SuperFlex loss (Eq.~\eqref{eq:superflex-loss}) using these queries.

\paragraph{Primitive budget.}
With few primitives, the network tends to spend them on the large parts
of the body. We double the number of part tokens from 16 to 32 and
remove the sparsity term $\mathcal{L}_{\mathrm{SPS}}$, which pushes the
network toward few primitives. The existence scores still switch off
unneeded primitives.

\subsection{Temporally Consistent Propagation}
\label{sec:key-frame-to-sequence}

We make the decomposition temporally consistent by computing it once, on
a keyframe, and then moving each primitive rigidly through the sequence.
Every primitive thus covers the same part and keeps its shape in all
frames, and only its pose changes.

\paragraph{Keyframe decomposition.}
\label{sec:TTO}
We run the network on the keyframe (Fig.~\ref{fig:our-method}, top
center) and optionally refine its prediction with TTO
(Sec.~\ref{sec:prelim-ff}), using the occupancy queries described above.
Since only the keyframe is refined, TTO runs once per sequence rather
than once per frame.

\paragraph{Rigid propagation.}
To move a primitive, we first use the point-to-part assignments to find
the points it covers: we assign each keyframe point $n$ to the primitive
with the highest assignment score, $\hat{k}_n=\arg\max_k A_{nk}$, and
denote the points of primitive $k$ by $\mathcal{I}_k=\{n:\hat{k}_n=k\}$.
After TTO, we keep the assignments of the feed-forward prediction. The
input meshes share their vertices across frames, so we know where each
keyframe point $\mathbf{x}_n^{(i)}$ lies in every target frame $j$, at
$\mathbf{x}_n^{(j)}$. For each
primitive and target frame, we find the rotation and translation that
best align its points (Fig.~\ref{fig:our-method}, bottom left),
\begin{equation}
\begin{aligned}
&(\Delta\mathbf{R}_k^{i\to j}, \Delta\mathbf{t}_k^{i\to j}) \\
&\quad= \underset{\mathbf{R}\in\mathrm{SO}(3),\,\mathbf{t}\in\mathbb{R}^3}{\arg\min}
\sum_{n\in\mathcal{I}_k}
\bigl\| \mathbf{R}\mathbf{x}_n^{(i)} + \mathbf{t} - \mathbf{x}_n^{(j)} \bigr\|_2^2,
\end{aligned}
\end{equation}
which the Kabsch algorithm solves in closed
form~\cite{kabsch1976solution}. Applying this rigid transform to the
keyframe pose of the primitive gives its pose in frame $j$
(Fig.~\ref{fig:our-method}, bottom right),
\begin{equation}
\begin{aligned}
\mathbf{R}_k^{(j)} &= \Delta\mathbf{R}_k^{i\to j}\mathbf{R}_k^{(i)}, \\
\mathbf{t}_k^{(j)} &= \Delta\mathbf{R}_k^{i\to j}\mathbf{t}_k^{(i)} + \Delta\mathbf{t}_k^{i\to j},
\end{aligned}
\end{equation}
while its shape, scale, tapering, and bending stay fixed. We compute
each frame directly from the keyframe, so errors do not accumulate
along the sequence. When point correspondences are noisy or unavailable,
a Huber-weighted solver~\cite{huber1992robust} or per-primitive ICP can
be used instead (Sec.~\ref{sec:supp_correspondence}).

\begin{figure*}[t]
\centering
\newlength{\qualcorrH}\setlength{\qualcorrH}{2.3in}
\begin{subfigure}[c]{0.60\textwidth}
  \centering
  \includegraphics[height=\qualcorrH,width=\linewidth,keepaspectratio]{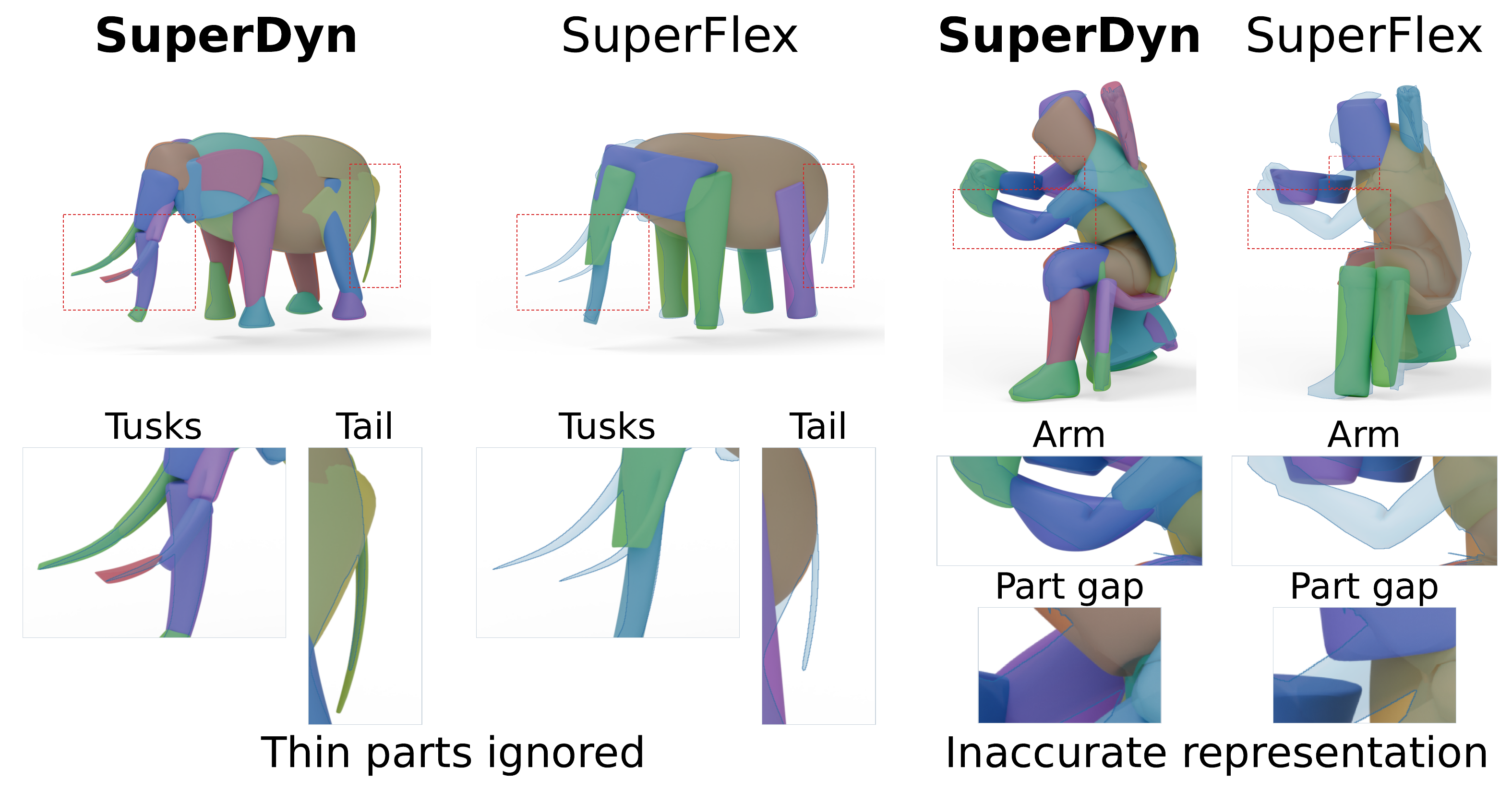}
  \caption{Thin-part preservation.}
  \label{fig:overlay}
\end{subfigure}\hfill
\begin{subfigure}[c]{0.38\textwidth}
  \centering
  \includegraphics[height=\qualcorrH,width=\linewidth,keepaspectratio]{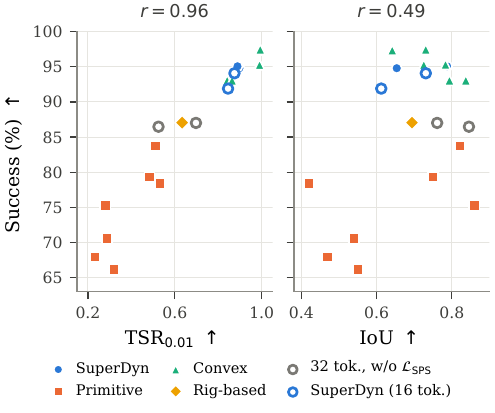}
  \caption{Planning success vs.\ TSR and IoU.}
  \label{fig:tsr-iou}
\end{subfigure}
\caption{\textbf{Thin-part coverage predicts planning success.}
(a) Decompositions with the ground-truth mesh overlaid in translucent blue; red boxes mark the enlarged regions. \methodname{} captures the elephant's tusks and tail and the character's arm, which SuperFlex misses or leaves with gaps.
(b) Planning success rate plotted against thin-structure recall (left) and volumetric IoU (right) on DT4D. Each point is one method; hollow markers show ablation variants (Table~\ref{tab:ablation-cumulative}). TSR is strongly correlated with planning success (Pearson $r = 0.96$), whereas IoU is only moderately correlated ($r = 0.49$).}
\label{fig:qual-and-corr}
\end{figure*}
\section{Thin-Structure Recall }
\label{sec:tsr}

Thin parts are critical for collision checking. A planner treats the
proxy as the obstacle, so if the proxy missed a tail or a leg, a planned
path can pass through it. Thin parts also tend to move fastest, so this
risk is highest for deforming objects. Volumetric IoU hardly reflects
such misses, since a tail holds only a small fraction of the object's
volume. We therefore introduce Thin-Structure Recall (TSR), the fraction
of an object's thin parts that a proxy covers.

\paragraph{Definition.}
Let $G$ be the voxelized solid of the full geometry, and let $P$ be the
voxelized union of the predicted primitives, that is, the space the
proxy occupies. Let $B_r$ be a ball of radius $r=0.01D$, where $D$ is
the diagonal of the bounding box of $G$. The morphological opening
$G\circ B_r$ is the union of all balls of radius $r$ that fit inside
$G$. A point of $G$ outside this union lies in a part too narrow to
contain such a ball, such as a tail or a tusk. We therefore define the
thin parts and TSR as
\begin{equation}
T_r = G \setminus (G \circ B_r), \qquad
\mathrm{TSR}_r = \frac{|T_r \cap P|}{|T_r|}.
\label{eq:tsr}
\end{equation}
With this radius, $T_r$ contains the parts thinner than about $2\%$ of
the object's size. We use 256 voxels along the longest side of the
bounding box and set $\mathrm{TSR}_r=1$ when the object has no thin
parts.

\paragraph{Excess volume.}
TSR alone can be raised by covering more space: a single box around the
whole object reaches $\mathrm{TSR}_r=1$. We therefore always report it
together with the volume the proxy adds outside the object, relative to
the object's volume,
\begin{equation}
\mathrm{ExcessVol./GT} = \frac{|P \setminus G|}{|G|}.
\end{equation}

\paragraph{TSR predicts planning success.}
In our planning experiments, methods with higher TSR achieve higher
planning success (Pearson $r=0.96$), whereas IoU is only moderately
correlated with it ($r=0.49$; Sec.~\ref{sec:motion-planning} and
Fig.~\ref{fig:tsr-iou}). TSR is therefore a much better indicator of
planning reliability than IoU.

\section{Experiments}
\label{sec:experiments}

\subsection{Experimental Setup}
\label{sec:experimental-setup}

\paragraph{Datasets and splits.}
We train on DeformingThings4D~\cite{li20214dcomplete} (DT4D) and evaluate on DT4D and Objaverse~\cite{deitke2023objaverse}. We split DT4D by character, so all 44 test characters (200 clips) are unseen during training; Sec.~\ref{sm:splits} lists the full splits. For DT4D evaluation, we retain up to 64 frames per test clip, yielding 9,981 frames. We use Objaverse only for evaluation, with 100 animated characters and 64 frames per character. For \methodname, \methodname{} (SuperDec), and SuperFlex+Kabsch, the keyframe is the first retained frame of each sequence.

\paragraph{Baselines.}
We compare with the primitive-based methods CSA~\cite{yang2021unsupervised}, SQ~\cite{paschalidou2019superquadrics}, SuperDec~\cite{fedele2025superdec}, SuperFlex~\cite{tavernini2026superflex}, EMS~\cite{liu2022robust}, and Marching Primitives~\cite{liu2023marching}, the convex methods V-HACD~\cite{mamou2016volumetric}, CoACD~\cite{wei2022approximate}, and Thul et al.~\cite{thul2018approximate}, and a rig-driven capsule baseline. For EMS and the convex methods, $K$ denotes the maximum number of parts. We use the project implementation of the sequence-aware method of Thul et al., which fits a shared partition using multiple keyframes and constructs corresponding deforming hulls. V-HACD and CoACD decompose frames independently, as do the primitive baselines unless explicitly marked \emph{+Kabsch}. SuperFlex+Kabsch predicts a keyframe decomposition and propagates it with the same per-primitive rigid alignment as our method. SuperFlex+TTO refines each frame independently, whereas SuperFlex+Kabsch+TTO refines only the keyframe and propagates that result. All reported SuperFlex variants use weights fine-tuned on DT4D, and CSA and SuperDec are trained from scratch on DT4D.

\emph{Skinning Capsules} follows the game-engine practice of attaching a capsule to each bone and moving it with the bone, using $K = 32$ capsules to match our primitive budget. It is fitted to 8 frames of the evaluated sequence and thus sees more of the motion than \methodname, which observes only the keyframe; details are in Sec.~\ref{sm:capsules}. We do not compare with animated sphere-meshes~\cite{thiery2016animated}, which have no public implementation.

\paragraph{Evaluation protocol.}
We average frame-level metrics within each sequence and weight all sequences equally. Within each dataset, all methods use the same reference geometry, normalization, and metric definitions. On DT4D, all methods are evaluated on the same 200 sequences. We use Objaverse to evaluate cross-dataset generalization after training on DT4D. Some methods fail on a few Objaverse characters, so we average each method over its valid sequences. Table~\ref{tab:sm-objaverse-clips} lists these counts, and Sec.~\ref{sm:implementation} gives our implementation details.

\subsection{Geometric Reconstruction and Thin-Part Coverage}
\label{sec:geometric-reconstruction}

\paragraph{Metrics.}
We measure thin-structure coverage with TSR and excess volume (Sec.~\ref{sec:tsr}). To capture large geometric misses, such as missing extremities, we also report the directed Hausdorff distance from the reference surface to the predicted surface, HD$_{\mathrm{GT}\to\mathrm{Pred}}$, and the symmetric HD, which captures large errors in either direction. Surface distances are normalized by $D$. Volumetric IoU, F-score, Chamfer distance, and the number of connected components are reported in Sec.~\ref{sm:metrics}.

\begin{figure}[t]
\centering
\includegraphics[width=0.85\linewidth]{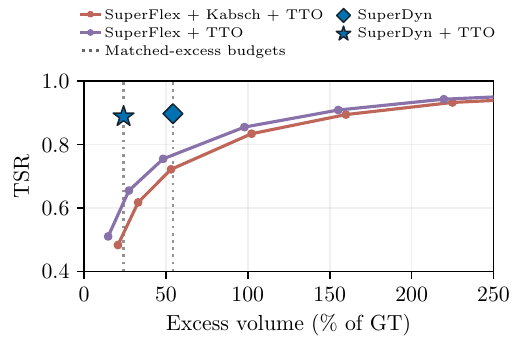}
\caption{\textbf{TSR at matched excess volume} on DT4D. SuperFlex
predictions are progressively dilated, while ours are not. At matched excess volume,
\methodname{} reaches higher TSR both with and without TTO.}
\label{fig:pareto}
\end{figure}

\paragraph{Coverage and reconstruction trade-off.}
On DT4D, \methodname{} roughly doubles TSR relative to SuperFlex (0.90
vs.\ 0.44; Table~\ref{tab:dt4d}). Qualitative examples appear in
Figs.~\ref{fig:teaser} and~\ref{fig:overlay}, with further comparisons
over full sequences in Fig.~\ref{fig:qualitative}. The higher
coverage comes at the cost of a larger proxy with higher excess volume.
The added volume alone, however, does not explain the gain. When SuperFlex+Kabsch+TTO is dilated to the
excess volume of \methodname, it reaches a TSR of only about 0.72, and at
the excess volume of \methodname+TTO about 0.52, compared with 0.90 and
0.89 for our variants (Fig.~\ref{fig:pareto}). TTO largely closes the
excess-volume gap: \methodname+TTO retains its TSR while reaching an
excess volume comparable to SuperFlex+Kabsch+TTO. The same trend holds on
Objaverse (Sec.~\ref{sm:objaverse}), where \methodname{} reaches a TSR of
0.91 compared with 0.43 for SuperFlex, and \methodname+TTO improves over
SuperFlex+TTO in both TSR (0.84 vs.\ 0.54) and excess volume (65\% vs.\
70\%).

\methodname{} also achieves the lowest HD and HD$_{\mathrm{GT}\to\mathrm{pred}}$ among primitive-based methods on DT4D, both with and without TTO, and the lowest among all methods on Objaverse after TTO, indicating fewer missing extremities.
\begin{table}[t]
\centering
\caption{Geometric reconstruction on DT4D. Top: primitive-based methods;
bottom: TTO variants, convex decompositions, and Skinning Capsules.
\textsc{SuperDyn}$_{\text{SuperDec}}$ applies our method to SuperDec,
+Kabsch denotes per-primitive Kabsch propagation, and numbers in
parentheses give $K$. HD is in units of $10^{-2}D$; size (amortized over
the clip) and time are per frame. Bold, underline, and italics mark the
first, second, and third place within each part.}
\label{tab:dt4d}
\sisetup{round-mode=places,round-precision=2,round-pad=false,
         group-digits=none, mode=text,
         reset-text-series=false, reset-text-shape=false}
\footnotesize
\setlength{\tabcolsep}{3pt}
\renewcommand{\arraystretch}{1.02}
\resizebox{\columnwidth}{!}{\begin{tabular}{@{}l|*{7}{r}@{}}
\toprule
Method & TSR$_{0.01}$ & Excess & HD & HD$_{\mathrm{GT}\to\mathrm{p}}$ & Parts & Size & Time \\
 & $\uparrow$ & (\%)\,$\downarrow$ & $\downarrow$ & $\downarrow$ & & (KiB)\,$\downarrow$ & (ms)\,$\downarrow$ \\
\midrule
\multicolumn{8}{@{}l}{\textit{Primitive-based}} \\
\midrule
\methodname & \textbf{\num{0.8971453907587048}} & \num{54.288191459010996} & \textbf{\num{7.026381294220939}} & \textbf{\num{6.353405358089645}} & \num{31.245} & \num{1.5458} & \textbf{\num{1.7629091803075287}} \\
\textsc{SuperDyn}$_{\text{SuperDec}}$ & \underline{\num{0.8714978260645435}} & \num{65.38718505619187} & \underline{\num{7.235980096100496}} & \underline{\num{6.6285009247410835}} & \num{31.79} & \num{1.5219} & \underline{\num{1.78940761272467}} \\
\midrule
SuperFlex & \num{0.43799667826138117} & \num{30.86790846058935} & \textit{\num{8.823840472334458}} & \num{8.807237174327314} & \num{8.911187331770932} & \num{1.7656} & \num{803.3042250852752} \\
SuperFlex+Kabsch & \num{0.4403029887535068} & \num{33.17590862397228} & \num{9.540749644348342} & \num{9.070707047738406} & \num{9.11} & \textit{\num{0.9573}} & \num{19.888404534676997} \\
SuperDec & \num{0.2928415608283258} & \textit{\num{23.82869818032273}} & \num{9.335542946248047} & \textit{\num{8.658981123170916}} & \num{12.784555544873932} & \num{1.7656} & \num{11.002657937471447} \\
SQ & \textit{\num{0.5285386476096218}} & \num{126.30840260246474} & \num{12.896447894699536} & \num{9.03426632032151} & \num{10.0} & \num{2.2070} & \textit{\num{2.4831376579673994}} \\
CSA & \num{0.22687884528362987} & \underline{\num{16.4325766425245}} & \num{10.605620575351274} & \num{10.13313231211047} & \num{10.821081133908713} & \num{2.6708} & \num{7.580673708977676} \\
EMS (8) & \num{0.3220739369365026} & \num{72.94689316914753} & \num{13.521189852646417} & \num{11.286720147106847} & \num{6.3930468850906985} & \textbf{\num{0.7555}} & \num{231.60816205972765} \\
EMS (32) & \num{0.3242262750910727} & \num{72.96828267006308} & \num{13.455904786229953} & \num{11.194427054509337} & \num{6.504278759620079} & \underline{\num{0.8453}} & \num{225.5146430165713} \\
Marching Prim. & \num{0.2843438157796177} & \textbf{\num{8.892105270443746}} & \num{11.517637532713964} & \num{11.508512288925601} & \num{17.050423608676464} & \num{1.6817} & 76.9e3 \\
\midrule
\multicolumn{8}{@{}l}{\textit{With TTO, convex, and rig-based}} \\
\midrule
\textsc{SuperDyn}+TTO & \num{0.8882563363692739} & \num{24.175314483103925} & \underline{\num{6.652076406669398}} & \num{5.016949488272704} & \num{31.245} & \textit{\num{1.5458}} & \textit{\num{203.61417627681948}} \\
\textsc{SuperDyn}$_{\text{SuperDec}}$+TTO & \num{0.8677} & \num{28.61} & \textit{\num{6.67}} & \num{5.35} & \num{31.79} & \underline{\num{1.5219}} & \textbf{\num{76.5}} \\
SuperFlex+TTO & \num{0.5106004416689539} & \textbf{\num{14.711175271587498}} & \num{8.095347011273063} & \num{7.857236120177973} & \num{8.911187331770932} & \num{1.7656} & 3.66e3 \\
SuperFlex+Kabsch+TTO & \num{0.48344477344926345} & \textit{\num{20.68430700902619}} & \num{9.593152445549688} & \num{8.376971270331705} & \num{9.11} & \textbf{\num{0.9573}} & \underline{\num{87.79910968308118}} \\
\midrule
Thul et al.\ (8) & \textit{\num{0.9910808438188254}} & \num{64.60527589737073} & \num{9.302728421183534} & \num{7.063875963026281} & \num{8.0} & \num{86.3419} & 4.51e3 \\
Thul et al.\ (16) & \num{0.9899344606222675} & \num{39.6765207420765} & \num{7.54793949799371} & \num{5.673999631984608} & \num{16.0} & \num{125.6080} & 5.11e3 \\
Thul et al.\ (32) & \num{0.9884877155221252} & \num{27.267630263602005} & \textbf{\num{6.598250409453424}} & \textit{\num{4.911818790472047}} & \num{31.984921875} & \num{164.6565} & 6.59e3 \\
V-HACD (8) & \num{0.8795905553111126} & \num{38.70493743845997} & \num{7.32451166615979} & \num{6.793098307510793} & \num{8.0} & \num{14.2358} & 1.29e3 \\
V-HACD (16) & \num{0.863183918092205} & \num{23.260994791254607} & \num{7.04404408474495} & \num{5.423973075109721} & \num{16.0} & \num{24.6690} & 1.58e3 \\
V-HACD (32) & \num{0.8409328297291355} & \underline{\num{15.375555308127087}} & \num{7.641240254969574} & \textbf{\num{4.317554185957258}} & \num{32.0} & \num{37.9735} & 1.43e3 \\
CoACD (8) & \textbf{\num{0.9944737520183312}} & \num{59.30304543705633} & \num{8.201378151821112} & \num{6.122571542040962} & \num{7.89445717254933} & \num{128.8693} & 64.4e3 \\
CoACD (16) & \underline{\num{0.9932147328374544}} & \num{37.99296182272602} & \num{7.977479504195471} & \underline{\num{4.505038622116676}} & \num{15.765245653592439} & \num{187.5780} & 99.3e3 \\
\midrule
Skinning Capsules & \num{0.6334661239188523} & \num{44.92640236124065} & \num{11.254970041918384} & \num{6.974993572279661} & \num{30.975} & \num{2.9488} & 23.4e3 \\
\bottomrule
\end{tabular}}
\vspace{-4.5mm}
\end{table}

\paragraph{Temporal stability.}
Since every frame is computed from the keyframe, we test whether accuracy degrades over time (Sec.~\ref{sec:temporal-stability}). The TSR of \methodname{} stays within about 0.04\, of its keyframe value, and its excess volume grows by about 7\%, as for SuperFlex+Kabsch. The TSR gain of TTO fades within the first frames, so the overall TSR is similar with and without TTO (Table~\ref{tab:dt4d}).
 
\paragraph{Comparison with convex and rig-based proxies.}
 CoACD and Thul et al. reach higher TSR, covering nearly all thin regions, but at a substantially higher storage
and construction cost. Skinning Capsules reaches a TSR of only
0.63 on DT4D (Table~\ref{tab:dt4d}), even though it relies on a
rig, uses a similar number of parts, and is fitted to multiple frames of
the evaluated sequence.

\subsection{Motion Planning around Deforming Obstacles}
\label{sec:motion-planning}
\paragraph{Setup and metrics.}
We evaluate planning in four synthetic rooms containing one or two
deforming obstacles (Fig.~\ref{fig:robot_navigation_rooms} shows the room layouts).
We plan with ST-RRT*~\cite{grothe2022st} for a spherical robot of radius
0.25\,m that moves in the horizontal plane at up to 3\,m/s. We keep only
room--sequence pairs for which the planner finds a collision-free path
around the original mesh, yielding 737 pairs and, with three planner
seeds each, 2,211 scenarios per method. A run fails if the planner
returns no path (e.g., the proxy blocks every passage or the search
times out) or if the returned path collides with the original mesh
sequence; Invalid GT counts only the latter. We also report the mean
search time of successful runs and the representation size.

\paragraph{Results.}
Table~\ref{tab:planning}
summarizes the results. Among primitive-based methods, \methodname{}
raises the success rate from 83.7\% for the strongest baseline to
95.1\%, and returns about four times fewer paths that collide with the
true geometry. Applied to SuperDec, our method raises its success rate
from 70.6\% to 93.8\% without TTO and reduces its invalid paths from
633 to 86. Skinning Capsules also falls short (87.1\%). A uniform safety margin does not close this gap: with the best margin, SuperFlex+TTO reaches only 91.5\% (Sec.~\ref{sm:safety-margin}). Compared with
convex decompositions, \methodname{} reaches a similar success rate
while requiring 8--56$\times$ less storage. CoACD still returns fewer invalid paths.

\begin{table}[t]
\centering
\caption{Robot path planning over 2,211 scenarios per method. Top:
feed-forward variants of our method and of fast primitive-based methods;
bottom: TTO variants, convex decompositions, and Skinning Capsules.
Notation and ranking follow Table~\ref{tab:dt4d}. Time is the mean search
time of successful runs, which rises for methods that also solve harder
scenarios. Invalid GT counts returned paths that are collision-free for
the proxy but collide with the reference geometry, and Thin cont.\ counts
those whose first contact lies in $T_r$. Ablation rows refer to
Table~\ref{tab:ablation-cumulative} (Abl.\ \#5 uses 16 part tokens) and
are not ranked.}
\label{tab:planning}
\footnotesize
\setlength{\tabcolsep}{3pt}
\renewcommand{\arraystretch}{1.02}
\resizebox{\columnwidth}{!}{\begin{tabular}{@{}l|rrrrr@{}}
\toprule
Method & Succ.\ rate & Time & Size & Invalid & Thin \\
 & (\%)\,$\uparrow$ & (s)\,$\downarrow$ & (KiB)\,$\downarrow$ & GT\,$\downarrow$ & cont.\,$\downarrow$ \\
\midrule
\multicolumn{6}{@{}l}{\textit{Feed-forward}} \\
\midrule
\methodname & \textbf{94.80} & 24.93 & 54.06 & \textbf{84} & \textbf{69} \\
\textsc{SuperDyn}$_{\text{SuperDec}}$ & \underline{93.80} & 20.47 & 53.91 & \underline{86} & \underline{72} \\
\midrule
SuperDec & 70.60 & 18.41 & \textit{34.25} & 633 & 356 \\
SQ & \textit{78.38} & 20.15 & \underline{26.93} & \textit{399} & \textit{273} \\
CSA & 67.98 & \underline{17.72} & 54.48 & 681 & 428 \\
EMS (32) & 66.21 & \textit{17.77} & \textbf{17.79} & 736 & 426 \\
Marching Prim. & 75.31 & \textbf{17.43} & 45.60 & 542 & 448 \\
\midrule
Abl.\ \#1 & 87.02 & 25.00 & 54.25 & 272 & 205 \\
Abl.\ \#5 & 91.90 & 22.83 & 27.90 & 129 & 106 \\
\midrule
\multicolumn{6}{@{}l}{\textit{With TTO, convex, and rig-based}} \\
\midrule
\textsc{SuperDyn}+TTO & 95.07 & 23.96 & 54.06 & 84 & 66 \\
\textsc{SuperDyn}$_{\text{SuperDec}}$+TTO & 94.93 & 21.09 & \textit{53.91} & 86 & 63 \\
\midrule
SuperFlex+TTO & 83.72 & 19.56 & \underline{38.45} & 355 & 284 \\
SuperFlex+Kabsch+TTO & 79.33 & 20.00 & \textbf{16.27} & 431 & 321 \\
\midrule
V-HACD (16) & 93.08 & \underline{17.80} & 425.86 & 149 & 81 \\
V-HACD (32) & 93.08 & \textbf{17.71} & 659.13 & 149 & 83 \\
Thul et al.\ (16) & 95.30 & 18.16 & 1803.10 & \textit{83} & \textit{58} \\
Thul et al.\ (32) & \textit{95.34} & \textit{18.02} & 2633.95 & 87 & 60 \\
CoACD (8) & \underline{97.38} & 18.54 & 2069.57 & \textbf{35} & \textbf{30} \\
CoACD (16) & \textbf{97.47} & 18.28 & 3037.18 & \underline{44} & \underline{33} \\
\midrule
Skinning Capsules & 87.06 & 19.41 & 145.24 & 262 & 194 \\
\midrule
Abl.\ \#1+TTO & 86.48 & 25.34 & 54.25 & 293 & 200 \\
Abl.\ \#5+TTO & 94.08 & 22.32 & 27.90 & 95 & 75 \\
\bottomrule
\end{tabular}}
\end{table}

\paragraph{Where do invalid paths fail?}
For each invalid path, we locate its first contact with the reference
geometry and test whether it lies in the thin-structure mask $T_r$
(Eq.~\eqref{eq:tsr}). Across all methods, 54--86\% of these contacts lie
in thin regions, which cover only about 19\% of the reference surface, a
3--4.5$\times$ over-representation. \methodname reduces such contacts
from 284 for SuperFlex+TTO to 69, comparable to convex decompositions
(Thin cont., Table~\ref{tab:planning}).

\begin{table}[t]
\centering
\caption{Cumulative ablation on DT4D. Rows 0--3 each add one change to the
row above; rows 4--5 are separate variants. +TTO uses 400 keyframe
refinement steps; Exc.\ is the excess volume in \%.}
\label{tab:ablation-cumulative}
\sisetup{round-mode=places,round-precision=2,round-pad=true,
         group-digits=none, mode=text}
\footnotesize
\setlength{\tabcolsep}{2.5pt}
\begin{tabular*}{\columnwidth}{@{\extracolsep{\fill}}cl|cc|cc|c@{}}
\toprule
& & \multicolumn{2}{c|}{Feed-forward} & \multicolumn{2}{c|}{+TTO} & \\
\# & Variant & TSR$_{0.01}$\,$\uparrow$ & Exc.\,$\downarrow$ & TSR$_{0.01}$\,$\uparrow$ & Exc.\,$\downarrow$ & Parts \\
\midrule
0 & SuperFlex+Kabsch & \num{0.4403} & \num{33.18} & \num{0.4834} & \num{20.68} & \num{9.11} \\
1 & +32 tok., w/o $\mathcal{L}_{\mathrm{SPS}}$ & \num{0.6972} & \num{27.29} & \num{0.5247} & \num{7.20} & \num{31.39} \\
2 & +our occ.\ queries & \num{0.8965} & \num{53.92} & \num{0.7995} & \num{15.65} & \num{31.39} \\
3 & +region-bal.\ sampling & \num{0.8971} & \num{54.29} & \num{0.8883} & \num{24.18} & \num{31.25} \\
  & \quad(\textsc{SuperDyn}) & & & & & \\
\midrule
4 & row 1 + region-bal. & \num{0.6870} & \num{25.21} & \num{0.5568} & \num{7.31} & \num{31.50} \\
5 & \textsc{SuperDyn}, 16 tok. & \num{0.8449} & \num{64.35} & \num{0.8743} & \num{34.61} & \num{15.93} \\
\bottomrule
\end{tabular*}
\end{table}

\subsection{Ablation Studies}
\label{sec:ablation}

Table~\ref{tab:ablation-cumulative} starts from SuperFlex and adds one
component at a time. All variants share the same training schedule and
are propagated with per-primitive rigid alignment.

\paragraph{Capacity.}
Increasing the number of part tokens to 32 and removing $\mathcal{L}_{\mathrm{SPS}}$ (row 1) lets the model use about 31 primitives and raises the feed-forward TSR from 0.44 to 0.70. Most of this gain, however, is lost during TTO, which favors a tight fit over thin-part coverage. After refinement, row 1 attains the lowest excess volume of all variants and a higher IoU than our method (Table~\ref{tab:quality_dt4d}), yet its TSR drops to 0.52 and it succeeds in only 86.5\% of planning scenarios, compared with 95.1\% for \methodname+TTO (Table~\ref{tab:planning}). A tight volumetric fit is therefore not sufficient for reliable planning.

\paragraph{Occupancy queries.}
Interior-weighted occupancy supervision (row~2) raises the feed-forward
TSR to 0.90, at the cost of higher excess volume. Without region-balanced
sampling, however, TTO still reduces TSR to 0.80.

\paragraph{Region-balanced sampling.}
Adding region-balanced sampling (row~3) leaves the feed-forward result
nearly unchanged but preserves thin-structure coverage through
refinement: TSR after TTO increases from 0.80 to 0.89. Region-balanced
sampling alone, without our occupancy queries (row~4), has little
effect. The two components are therefore complementary: the occupancy
queries make the network cover thin parts, and the balanced input keeps
them represented when the keyframe is refined.

\paragraph{Primitive budget.}
With 16 part tokens, the budget of SuperFlex, \methodname still nearly
doubles TSR relative to SuperFlex (0.87 vs.\ 0.48 after TTO; row~5),
showing that the gain does not stem from additional capacity alone.
In planning, this variant reaches 94.1\% success at half the storage of
the 32-token model (Table~\ref{tab:planning}), showing that the
method remains effective under a tight primitive budget. With fewer
primitives, however, each must cover a larger region, which increases
excess volume; we therefore use 32 tokens in all other experiments.

\section{Conclusion}
\label{sec:conclusion}
We presented \methodname, a general approach that makes feed-forward superquadric decomposition networks cover thin parts and stay temporally consistent. Missing thin parts, which IoU barely registers, cause most invalid paths around deforming obstacles, and TSR captures them. With both SuperFlex and SuperDec, \methodname{} reaches planning success comparable to convex decomposition at a fraction of the storage. Future work could aggregate information from multiple frames instead of one keyframe. We discuss limitations, including the dependence on the keyframe pose and on point correspondences, in Sec.~\ref{sm:limitations}.

\section*{Acknowledgments}
This research  is supported by the Israel Science
Foundation through a personal grant (ISF 1425/26) and an equipment grant (ISF 4024/26). Yoni Kasten is a Dr. Natalie Shaffer Faculty Fellow.

\appendix

\twocolumn[{\centering
  {\LARGE\bfseries Supplementary Material\par}
  \vspace{1em}
}]

\setcounter{figure}{0}
\setcounter{table}{0}
\setcounter{equation}{0}

\renewcommand{\thesection}{S\arabic{section}}
\renewcommand{\thefigure}{S\arabic{figure}}
\renewcommand{\thetable}{S\arabic{table}}
\renewcommand{\theequation}{S\arabic{equation}}

\renewcommand{\theHsection}{supp.\arabic{section}}
\renewcommand{\theHfigure}{supp.\arabic{figure}}
\renewcommand{\theHtable}{supp.\arabic{table}}
\renewcommand{\theHequation}{supp.\arabic{equation}}

\renewcommand{\thesection}{S\arabic{section}}
\renewcommand{\thetable}{S\arabic{table}}
\renewcommand{\thefigure}{S\arabic{figure}}
\renewcommand{\theequation}{S\arabic{equation}}
\renewcommand{\topfraction}{0.95}
\renewcommand{\textfraction}{0.05}
\renewcommand{\floatpagefraction}{0.8}

\section*{Overview}

This supplementary material complements the main paper with implementation
details, additional results, and further analyses.
Sections~\ref{sm:splits}--\ref{sm:metrics} describe the data splits,
implementation, baselines, planning setup, and evaluation metrics.
Sections~\ref{sm:objaverse}--\ref{sm:limitations} report additional
experiments:
\begin{itemize}
  \item Sec.~\ref{sm:objaverse}: results on Objaverse~\cite{deitke2023objaverse} without fine-tuning.
  \item Sec.~\ref{sm:tsr-radius}: TSR at different radii and its relation to collisions within each method.
  \item Secs.~\ref{sec:temporal-stability} and~\ref{sec:keyframe-selection}: accuracy over time and the choice of keyframe.
    \item Sec.~\ref{sec:supp_correspondence}: robustness to jitter, outliers,
    and independent and correlated drift in the correspondences, a robust
    solver that tolerates 40\% wrong correspondences, and propagation
    without correspondences via ICP.
  \item Sec.~\ref{sm:limitations}: limitations.
\end{itemize}

\section{Dataset Splits}
\label{sm:splits}

Table~\ref{tab:sm-dt4d-splits} lists the DT4D~\cite{li20214dcomplete} splits. Test characters are
disjoint from the training characters. The validation split shares
characters, but no clips, with the test split, which is why the character
counts of the splits sum to more than the total. We train for a fixed
schedule and evaluate the final checkpoint, without checkpoint selection on
validation data. We used 20 validation clips to select the number of TTO iterations and fixed this budget at 400 for all SuperDyn test runs with TTO (Sec.~\ref{sec:tto-iterations}).

\begin{table}[h]
\centering
\small
\caption{DT4D splits. Frame counts are before subsampling.}
\label{tab:sm-dt4d-splits}
\begin{tabular}{@{}lrrr@{}}
\toprule
Split & Characters & Clips & Frames \\
\midrule
Train & 56 & 1,660 & 91,580 \\
Val   & 27 &    93 &  8,282 \\
Test  & 44 &   200 & 21,744 \\
\midrule
Total & 103 & 1,953 & 121,606 \\
\bottomrule
\end{tabular}
\end{table}

\section{Implementation Details}
\label{sm:implementation}

\paragraph{Training.}
\methodname uses $N=4{,}096$ input points from $L=32$ regions with 128
points each, and predicts $K=32$ candidate primitives. We initialize the
network from ShapeNet-pretrained~\cite{chang2015shapenet} SuperFlex~\cite{tavernini2026superflex} weights. Since the part queries
are zero vectors combined with sinusoidal positional encodings, we extend
them from 16 to 32 by adding positional encodings for the new indices,
which distinguishes the new queries without changing any pretrained
weights. We then fine-tune on DT4D for 350,000 Adam~\cite{kingma2014adam} updates with batch size
16. The initial learning rate is $3\times10^{-5}$ and is halved after
230,000 and 270,000 updates. We disable dropout and apply random yaw
augmentation. Each training example uses 4,096 occupancy queries: 3,072
outside the object, 512 inside it near the surface, and 512 throughout its
interior (Sec.~\ref{sec:thin-parts}).

\paragraph{\methodname (SuperDec).}
We implement \methodname (SuperDec)~\cite{fedele2025superdec} in the same way. We train \methodname (SuperDec) using the same fine-tuning protocol,
initializing it from the final checkpoint of the SuperDec
baseline trained from scratch on DT4D with its original training
objective. We expand its part queries from 16 to 32 using the same
positional-encoding extension described above, preserving all learned
weights, and fine-tune for an additional 350,000 updates with a fresh
Adam optimizer.

\paragraph{Test-time optimization.}
\methodname+TTO runs 400 Adam steps on the keyframe with the
per-parameter-group learning rates of SuperFlex (see Table~\ref{tab:tto_lr}) and 100,000 fixed occupancy queries in
the same proportions, while keeping the network weights and the
point-to-part assignments fixed. SuperFlex+TTO follows the original
procedure~\cite{tavernini2026superflex} with 1,000 steps per frame. We treat
primitives with existence scores greater than $0.5$ as active.

\paragraph{Point correspondences.}
\label{sm:correspondences}
The meshes of each sequence share their vertices across frames. Since the
input points are sampled on the surface rather than at vertices, we track
each sampled point through its triangle and barycentric coordinates, which
stay fixed throughout the sequence. These point correspondences are used for
rigid propagation (Sec.~\ref{sec:key-frame-to-sequence}).
In real captures, correspondences would have to be estimated and would
contain errors. Section~\ref{sec:supp_correspondence} evaluates propagation
under jitter, outliers, and drift, and shows that it also works without
any correspondences, by registering each primitive to independently
sampled point clouds with ICP.

\paragraph{Timing.}
Creation times are wall-clock per frame. GPU methods run on a single RTX PRO
6000 Blackwell, and CPU methods run on 16 CPU cores with exclusive access.

\begin{table}[t]
    \centering
    \caption{Adam learning rates for test-time optimization.
    Rates apply to the optimization variables of each parameter group.
    Tapering and bending are omitted in \methodname (SuperDec).}
    \label{tab:tto_lr}
    \begin{tabular}{lc}
        \hline
        Parameter group & Learning rate \\
        \hline
        Scale           & $2\times10^{-2}$ \\
        Shape exponents & $1\times10^{-2}$ \\
        Rotation        & $2\times10^{-3}$ \\
        Translation     & $2\times10^{-3}$ \\
        Tapering        & $7\times10^{-3}$ \\
        Bending         & $6\times10^{-3}$ \\
        \hline
    \end{tabular}
\end{table}

\section{Superquadric Deformations}
\label{sm:deformations}

Tapering and bending~\cite{barr1984global} are applied in the local coordinate system of a
primitive, with bending applied after tapering. Tapering scales the
cross-section linearly along the $z$-axis,
\begin{equation}
T_{\boldsymbol{\tau}}(x,y,z) =
\bigl(
x\,(1+\tau_x z/s_z),\;
y\,(1+\tau_y z/s_z),\;
z
\bigr).
\end{equation}
A bend along the $z$-axis has a curvature $\kappa$ and a direction angle
$\alpha$. With $r=x\cos\alpha+y\sin\alpha$, $\rho=\kappa^{-1}-r$, and
$\gamma=\kappa z$, the bending map is
\begin{equation}
B_{\kappa,\alpha}(x,y,z) =
\left(x+\delta\cos\alpha,\; y+\delta\sin\alpha,\; \rho\sin\gamma\right),
\end{equation}
where $\delta=\rho(1-\cos\gamma)$. The map reduces to the identity as
$\kappa\to 0$. Bends along the other two axes are defined by permuting the
local coordinates. Following SuperFlex, we apply the three bends
sequentially along the $y$-, $x$-, and $z$-axes after tapering, which
yields the local deformation
\begin{equation}
D =
B^{(z)}_{\kappa_z,\alpha_z}
\circ B^{(x)}_{\kappa_x,\alpha_x}
\circ B^{(y)}_{\kappa_y,\alpha_y}
\circ T_{\boldsymbol{\tau}}.
\end{equation}

\section{Number of TTO Iterations}
\label{sec:tto-iterations}

\begin{figure*}[t]
\centering
\includegraphics[width=\textwidth]{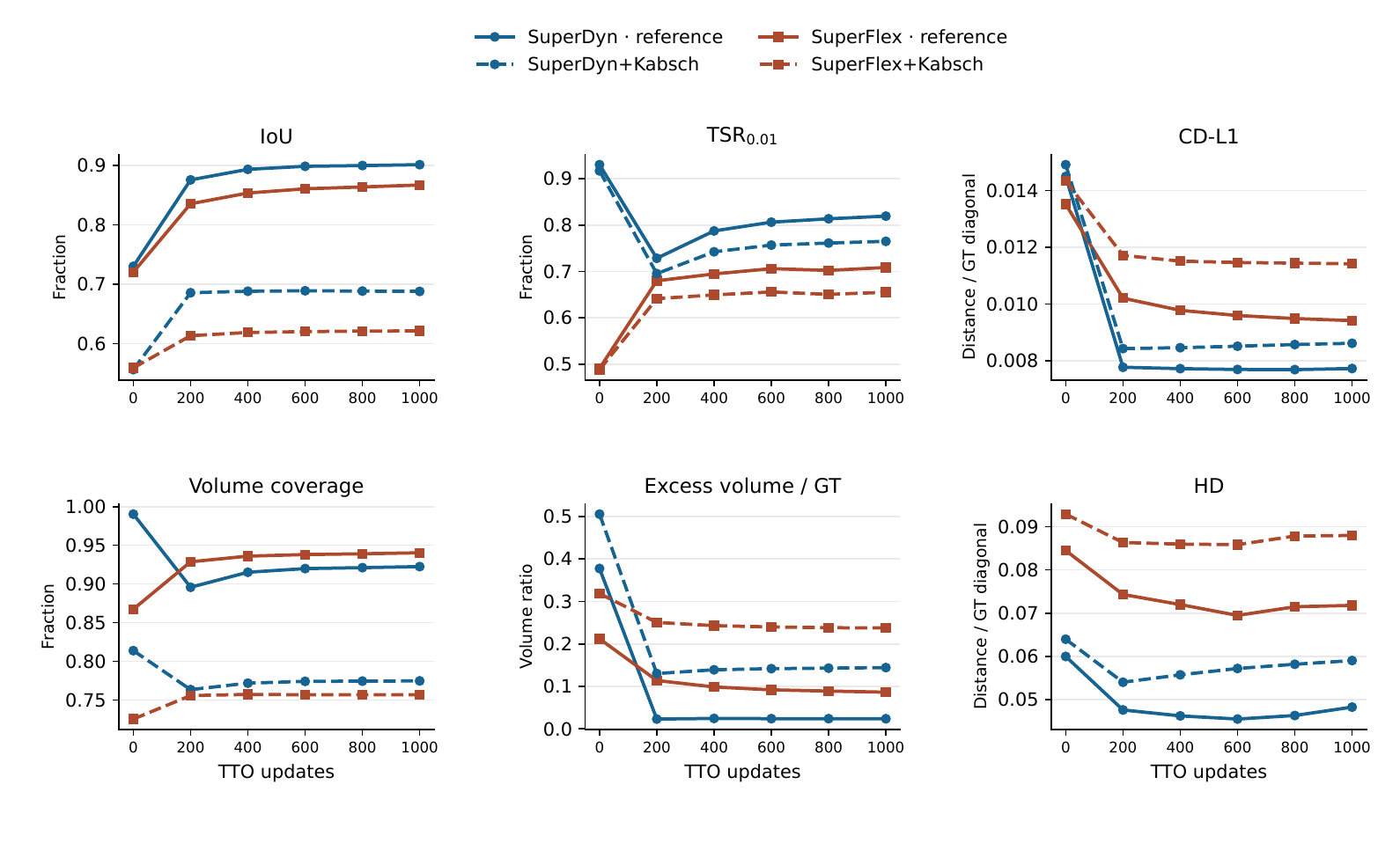}
\captionsetup{hypcap=false}
\captionof{figure}{\textbf{Effect of TTO iterations on 20 validation clips.} Solid curves show reference-frame results; dashed curves show clip averages with Kabsch propagation. At 400 iterations, SuperDyn achieves 95.4\% of its reference-frame IoU gain and matches its clip-averaged IoU at 1,000 iterations.}
\label{fig:tto_ablation}
\captionsetup{hypcap=true}
\end{figure*}

We select the number of TTO iterations on 20 validation clips
(Fig.~\ref{fig:tto_ablation}). For \methodname{}, most of the gain occurs within the first 200
updates: CD-L1 and excess volume on the reference frame reach a plateau,
and IoU rises from 0.73 to 0.88. At 400 updates, \methodname{} achieves
95.4\% of the reference-frame IoU gain obtained at 1,000 updates, and its
clip-averaged IoU with Kabsch propagation already matches the value at
1,000 updates. Further updates therefore refine only the keyframe, and this
refinement does not carry over to the propagated frames.
Each TTO update of \methodname{} is also more expensive than an update of
SuperFlex, partly because it optimizes 32 rather than 16 candidate
primitives. Even with fewer updates, \methodname{}+TTO requires 204\,ms per
frame on DT4D, compared with 88\,ms for SuperFlex+Kabsch+TTO with 1,000
updates (Table~\ref{tab:dt4d}). We therefore use 400 updates, which captures
nearly all of the useful gain while keeping the cost of TTO bounded. For
SuperFlex+TTO, we keep the 1,000 updates of the original procedure.

\section{Skinning Capsules Baseline}
\label{sm:capsules}

\paragraph{Rig estimation.}
DT4D sequences do not provide skeletons, so we estimate a rig from each
sequence with Smooth Skinning Decomposition with Rigid Bones
(SSDR)~\cite{le2012smooth}, using the Dem Bones implementation. With the
first frame as the reference pose, SSDR fits a single set of skinning
weights, with at most four bones per vertex, and a rigid transformation
for each bone in every retained frame. We request up to 32 bones and fit
one capsule per bone, matching the primitive budget of \methodname. The rig is estimated from all frames of the evaluated sequence, while the capsule shapes are fitted on eight of them (Sec.~\ref{sec:experimental-setup}). This baseline therefore observes more of the motion than \methodname, which uses only the keyframe.

\paragraph{Capsule fitting.}
Each capsule is initialized from the reference-pose vertices of its bone:
its axis is the leading eigenvector of their skinning-weighted covariance,
and its length and radius are set from robust percentiles of the vertex
positions along and around this axis. We then refine the capsule centers,
lengths, and radii, with the axes kept fixed relative to their bones, on
eight evenly spaced frames of the sequence. In each frame, we transform
the capsules by their bone transformations and compare their union with
the occupancy of the original mesh, using queries sampled inside, outside,
and around the object and labeled by the generalized winding number. We
minimize a binary cross-entropy loss on a soft union of the capsules,
which penalizes both missing and excess volume, using Adam~\cite{kingma2014adam} for 1,200
steps with a small penalty on center displacement. 

\paragraph{Evaluation.}
In every frame, each capsule follows the rigid transformation of its
bone, and the proxy is the union of the transformed capsules. The capsule
shapes are fitted once per sequence and are not refitted per frame.

\section{Planning Details}
\label{sm:planning}

\begin{figure}
\centering
\includegraphics[width=0.5\textwidth]{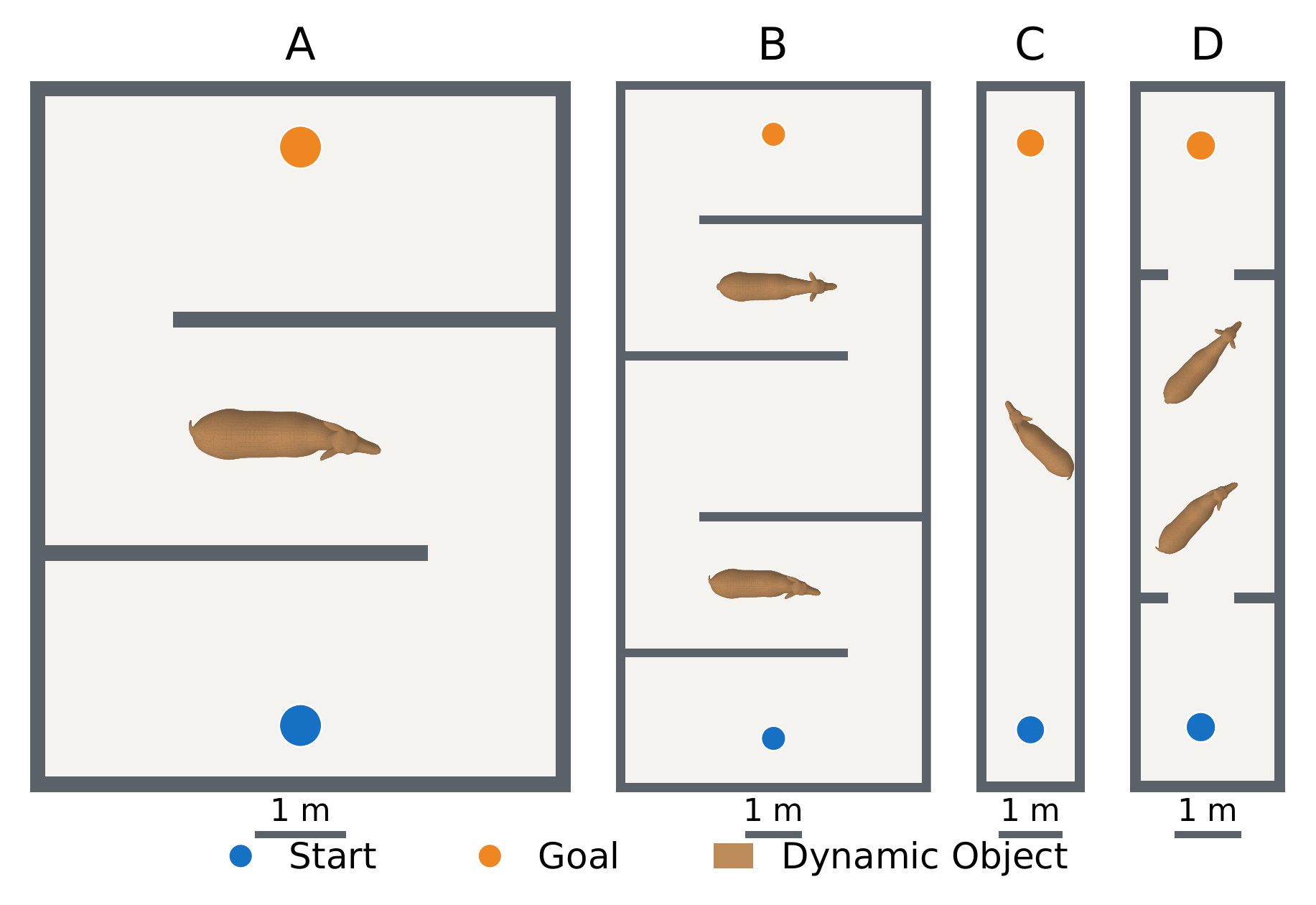}
\captionsetup{hypcap=false}
\captionof{figure}{Robot navigation environments. Top-down views of rooms A-D. Dynamic obstacles are shown using the ground-truth deer mesh at the initial simulation time. Blue and orange circles indicate the robot's start and goal positions, respectively. Rooms are displayed at different scales for visual alignment; each scale bar represents 1 m.}
\label{fig:robot_navigation_rooms}
\captionsetup{hypcap=true}
\end{figure}

\paragraph{Environments.}
We use four synthetic rooms (Fig.~\ref{fig:robot_navigation_rooms}), each
containing one or two deforming obstacles between the robot's start and
goal. Each obstacle replays the retained frames of a DT4D test sequence at
11.2\,frames/s, reversing playback at the sequence endpoints (ping-pong
playback). Obstacles have fixed world placements, and their deformation
comes from the original animation. Scaling and placement are computed from
the original mesh sequence, independently of the evaluated proxy, and are
shared by all methods. In rooms A and B, each sequence is uniformly scaled
so that its median largest extent is about 2.42\,m for elephant, cattle,
and deer sequences and 2.66\,m for all others, subject to a cap on its
longitudinal extent over the sequence. In rooms C and D, an additional
uniform scale fits the sequence within the room width with at least 1\,cm
clearance on either side; the obstacle orientation in C differs from that
in D by $90^\circ$. We retain only room--sequence pairs for which a
collision-free path around the original mesh sequence exists, as verified
by our planner and validation. This yields 196, 175, 175, and 191 pairs in
rooms A--D (737 in total) and, with three planner seeds per pair, 2{,}211
scenarios per method.

\paragraph{Planner.}
We use the OMPL~1.7.0~\cite{sucan2012open} implementation of
ST-RRT*~\cite{grothe2022st}. The robot is a sphere of radius 0.25\,m whose
center moves in the horizontal plane at a height of 0.5\,m and a maximum
speed of 3\,m/s. During planning, each obstacle is represented only by the
evaluated proxy, held constant between retained frames (zero-order hold).
Collision checks along each space--time edge are spaced by at most 2\,cm of
robot translation and 10\,ms of simulation time, and edges are split at
frame boundaries. A state is accepted only if the collision check is
definitely clear. Each search stops at the first solution and has a time
budget of 300\,s. Planning time includes all collision queries but excludes
geometry preparation, which is performed once beforehand.

\paragraph{Validation.}
Each returned path is validated offline against the original mesh
sequence, whose vertices are linearly interpolated between retained frames.
We sample the path in space and time at 1\,cm and 5\,ms and refine to
0.5\,cm and 2.5\,ms, and further to 0.25\,cm and 1.25\,ms if the result
has not stabilized. A sample collides if the robot center lies within
0.25\,m of the reference surface (with a 10\,$\mu$m tolerance) or inside the
reference solid, as determined by the generalized winding number. A run
succeeds only if its path passes this validation; runs whose validation
remains ambiguous within the tolerance are counted as failures
(Sec.~\ref{sec:supp_failures}).

\section{Additional Metrics}
\label{sm:metrics}

\begin{figure*}[t]
\centering
\includegraphics[width=\linewidth]{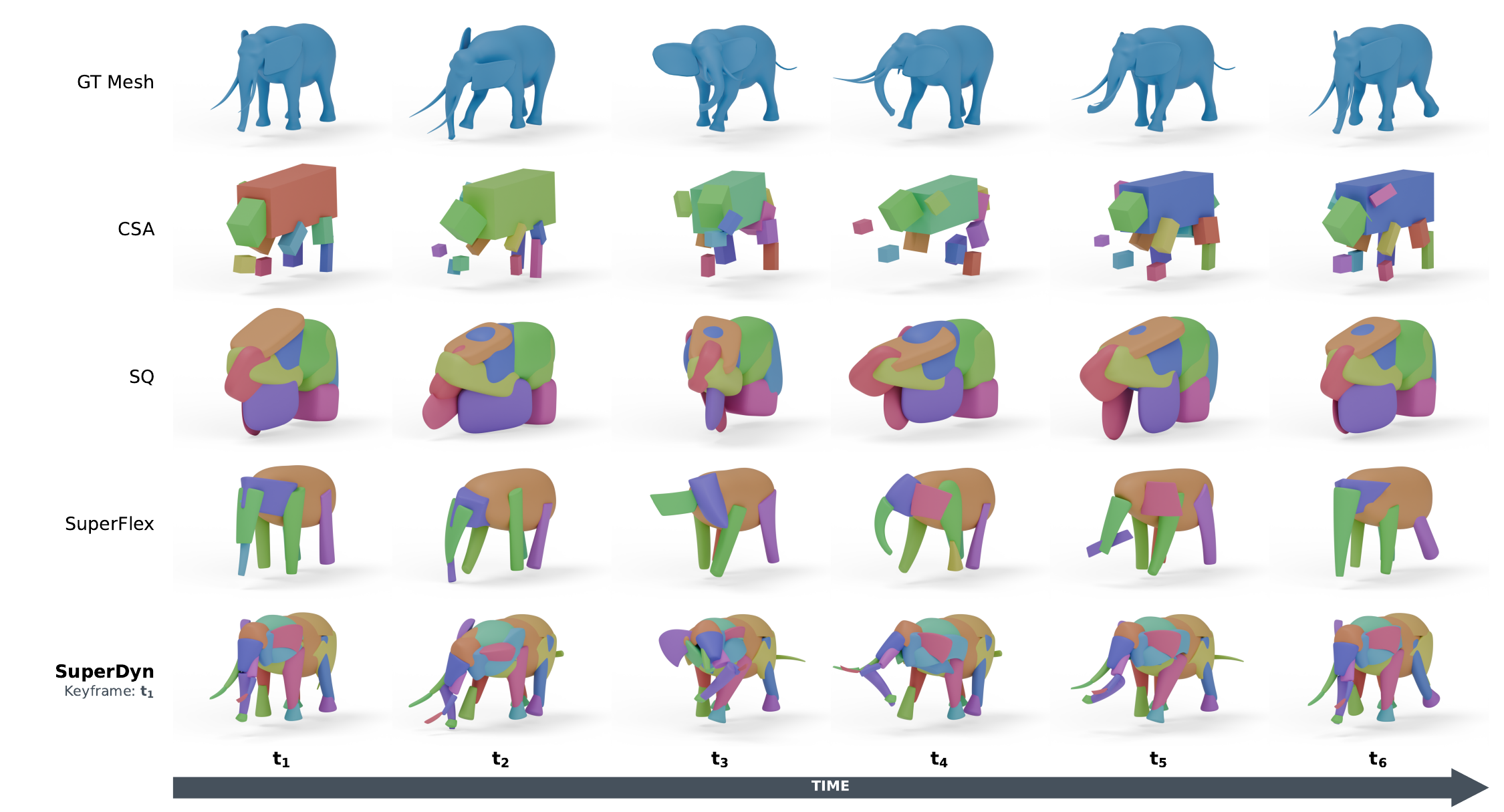}
\caption{Qualitative comparison on a deforming elephant sequence. Columns show six selected frames in chronological order. Rows show the ground-truth mesh (blue), CSA, SQ, SuperFlex+TTO, and SuperDyn+TTO. All rows use the same source frames, camera viewpoint, and scale. Colors do not indicate correspondence between methods.}
\label{fig:qualitative}
\end{figure*}

Given the clean reference meshes in our evaluation, we use the maximum
surface distance in the Hausdorff metrics to retain sensitivity to localized
reconstruction failures. We additionally report volumetric IoU, F-score at a
distance threshold of $1\%$ of $D$, and symmetric Chamfer distance (CD).
Surface distances are normalized by $D$, or by $D^2$ for squared CD. To
assess whether gaps between primitives fragment the reconstruction, we
report the number of connected components (NCC) in the predicted union.
Each evaluation sequence contains a single character whose geometry we
assume to be connected, so a well-connected reconstruction has an NCC of
one. Tables~\ref{tab:quality_dt4d} and~\ref{tab:quality_objaverse} report
these metrics for all methods on DT4D and Objaverse, respectively. \methodname{} has a lower IoU and F-score than SuperFlex, since its larger excess volume adds geometry outside the object. Planning success is more closely related to TSR than to IoU ($r=0.96$ vs.\ $0.49$; Fig.~\ref{fig:tsr-iou}). Among the primitive-based methods with TTO, \methodname{} produces the fewest connected components on both datasets.
Fig.~\ref{fig:qualitative} shows a full sequence compared with CSA, SQ, and SuperFlex+TTO.

\begin{table}[!tbp]
\centering
\footnotesize
\setlength{\tabcolsep}{4pt}
\begin{tabular}{@{}lcccc@{}}
\toprule
Method & IoU $\uparrow$ & F-score $\uparrow$ & CD-L1 $\downarrow$ & NCC $\downarrow$ \\
 & & & (\%$D$) & \\
\midrule
\multicolumn{5}{@{}l}{\textit{Primitive-based}} \\
\midrule
\textsc{SuperDyn} & 0.65 & 0.28 & 1.55 & \underline{1.27} \\
\textsc{SuperDyn}$_{\mathrm{SuperDec}}$ & 0.61 & 0.21 & 1.75 & 1.36 \\
\addlinespace[2pt]
SuperFlex & \underline{0.70} & \underline{0.42} & \underline{1.39} & 2.25 \\
SuperFlex+Kabsch & 0.68 & 0.40 & 1.51 & 2.41 \\
SuperDec & 0.54 & 0.31 & 1.74 & 6.56 \\
SQ & 0.42 & 0.18 & 2.86 & \textbf{1.10} \\
CSA & 0.47 & 0.28 & 1.95 & 7.22 \\
EMS (8) & 0.55 & 0.37 & 2.17 & 4.25 \\
EMS (32) & 0.55 & 0.37 & 2.16 & 4.28 \\
Marching Prim. & \textbf{0.86} & \textbf{0.73} & \textbf{0.99} & 2.12 \\
\midrule
\multicolumn{5}{@{}l}{\textit{With TTO, convex, and rig-based}} \\
\midrule
\textsc{SuperDyn}+TTO & 0.79 & 0.60 & 1.05 & 1.51 \\
\textsc{SuperDyn}$_{\mathrm{SuperDec}}$+TTO & 0.76 & 0.55 & 1.13 & 1.82 \\
SuperFlex+TTO & \underline{0.82} & \underline{0.64} & \textbf{1.01} & 1.78 \\
SuperFlex+Kabsch+TTO & 0.75 & 0.54 & 1.26 & 2.28 \\
\addlinespace[2pt]
Thul et al. (8) & 0.62 & 0.41 & 1.71 & \textbf{1.00} \\
Thul et al. (16) & 0.73 & 0.52 & 1.30 & \textbf{1.00} \\
Thul et al. (32) & 0.78 & 0.60 & 1.11 & \textbf{1.00} \\
V-HACD (8) & 0.71 & 0.50 & 1.35 & \underline{1.04} \\
V-HACD (16) & 0.79 & 0.60 & 1.09 & 1.06 \\
V-HACD (32) & \textbf{0.84} & \textbf{0.65} & \underline{1.03} & 1.10 \\
CoACD (8) & 0.64 & 0.36 & 1.62 & 4.55 \\
CoACD (16) & 0.73 & 0.45 & 1.35 & 7.97 \\
\addlinespace[2pt]
Skinning Capsules & 0.69 & 0.41 & 1.52 & 4.49 \\
\midrule
\multicolumn{5}{@{}l}{\textit{Ablations (not ranked)}} \\
\midrule
Abl.\ \#1 & 0.76 & 0.54 & 1.13 & 2.07 \\
Abl.\ \#1+TTO & 0.85 & 0.72 & 0.90 & 2.94 \\
Abl.\ \#2 & 0.66 & 0.28 & 1.55 & 1.19 \\
Abl.\ \#2+TTO & 0.83 & 0.67 & 0.97 & 1.71 \\
Abl.\ \#4 & 0.77 & 0.56 & 1.09 & 2.42 \\
Abl.\ \#4+TTO & 0.85 & 0.72 & 0.89 & 3.09 \\
Abl.\ \#5 & 0.61 & 0.23 & 1.76 & 1.31 \\
Abl.\ \#5+TTO & 0.73 & 0.48 & 1.24 & 1.45 \\
\bottomrule
\end{tabular}
\caption{DT4D reconstruction quality, averaged equally over clips. F-score uses a threshold of $0.01D$; CD-L1 is the symmetric mean unsquared surface distance, normalized by the GT bounding-box diagonal $D$ and reported as a percentage. Bold and underline indicate the best and second-best values within each of the two groups. Ablation numbering follows Table~\ref{tab:ablation-cumulative}; ablations~0 and~3 are omitted, since they coincide with SuperFlex+Kabsch and \textsc{SuperDyn}, respectively. }
\label{tab:quality_dt4d}
\end{table}

\begin{table}[!tbp]
\centering
\small
\setlength{\tabcolsep}{4pt}
\begin{tabular}{l|rrrr}
\toprule
Method & IoU $\uparrow$ & F-score $\uparrow$ & CD-L1 $\downarrow$ & NCC $\downarrow$ \\
 & & & (\%$D$) & \\
\midrule
\multicolumn{5}{l}{\textit{Primitive-based}} \\
\midrule
\textsc{SuperDyn} & 0.53 & 0.29 & 1.54 & \underline{1.50} \\
\textsc{SuperDyn}$_{\mathrm{SuperDec}}$ & 0.48 & 0.21 & 1.77 & 1.55 \\
\midrule
SuperFlex & \underline{0.58} & \underline{0.45} & \textbf{1.41} & 2.94 \\
SuperFlex+Kabsch & 0.57 & 0.44 & \underline{1.50} & 3.12 \\
SuperDec & 0.42 & 0.31 & 1.83 & 6.18 \\
SQ & 0.28 & 0.14 & 3.36 & \textbf{1.17} \\
CSA & 0.35 & 0.27 & 2.14 & 7.54 \\
EMS (8) & -- & -- & -- & -- \\
EMS (32) & 0.43 & 0.40 & 2.58 & 3.63 \\
Marching Prim. & \textbf{0.73} & \textbf{0.65} & 1.92 & 1.51 \\
\midrule
\multicolumn{5}{l}{\textit{With TTO, convex, and rig-based}} \\
\midrule
\textsc{SuperDyn}+TTO & \underline{0.75} & \textbf{0.74} & \textbf{0.86} & 1.99 \\
\twoline{\textsc{SuperDyn}$_{\mathrm{SuperDec}}$}{\,+TTO} & 0.70 & 0.70 & 0.92 & 2.84 \\
SuperFlex+TTO & 0.72 & 0.68 & 1.02 & 2.41 \\
\twoline{SuperFlex+Kabsch}{\,+TTO} & 0.68 & 0.63 & 1.17 & 2.64 \\
\midrule
Thul et al. (8) & -- & -- & -- & -- \\
Thul et al. (16) & 0.60 & 0.57 & 1.24 & \textbf{1.04} \\
Thul et al. (32) & 0.61 & 0.62 & 1.10 & \underline{1.04} \\
V-HACD (8) & -- & -- & -- & -- \\
V-HACD (16) & 0.71 & 0.68 & 0.95 & 1.08 \\
V-HACD (32) & \textbf{0.76} & \underline{0.73} & \underline{0.89} & 1.14 \\
CoACD (8) & -- & -- & -- & -- \\
CoACD (16) & 0.62 & 0.53 & 1.22 & 8.16 \\
\midrule
\twoline{Skinning}{Capsules} & 0.50 & 0.38 & 1.98 & 6.07 \\
\bottomrule
\end{tabular}
\caption{Objaverse reconstruction quality, averaged equally over clips.
  F-score uses a threshold of $0.01D$; CD-L1 is the symmetric mean
  unsquared surface distance normalized by the GT bounding-box diagonal
  $D$, reported as a percentage. Bold and underline indicate the best and
  second-best values within each of the two groups. Unavailable runs are
  marked --. Methods that fail on some characters, including
  \methodname{}, are averaged over their valid clips
  (Table~\ref{tab:sm-objaverse-clips}).}
\label{tab:quality_objaverse}
\end{table}

\section{Cross-Dataset Generalization}
\label{sm:objaverse}

We evaluate all models trained on DT4D, without fine-tuning, on 100
animated characters from Objaverse~\cite{deitke2023objaverse} with 64 frames each. These characters
differ from DT4D in topology, proportions, and level of detail. Some
methods fail on a few characters; we average each method over its valid
clips (Table~\ref{tab:sm-objaverse-clips}). \methodname{} could not process one character (Velociraptor)
because an FPS-anchor neighborhood contained fewer than the
128 distinct points required by our input sampling procedure.
The $K=8$ variants of EMS~\cite{liu2022robust}, Thul et al.~\cite{thul2018approximate}, V-HACD~\cite{mamou2016volumetric}, and CoACD~\cite{wei2022approximate}
were not evaluated on Objaverse.
For the convex decomposition methods, meshes with many connected
components incurred prohibitively long runtimes, preventing
practical completion of the runs.
Table~\ref{tab:objaverse} reports TSR, excess volume, and
Hausdorff distances, and Table~\ref{tab:quality_objaverse} reports IoU,
F-score, Chamfer distance, and connectivity.

\paragraph{Thin-part coverage transfers.}
The gain in TSR carries over almost unchanged. Without TTO, \methodname{}
reaches a TSR of 0.91, compared with 0.43 for SuperFlex, nearly the same
gap as on DT4D (0.90 vs.\ 0.44). Applied to SuperDec, it reaches 0.89,
compared with 0.33. With TTO, \methodname{} reaches 0.84, compared with
0.54 for SuperFlex+TTO.

\paragraph{Excess volume grows without TTO.}
On Objaverse, all primitive-based methods add more volume than on DT4D.
Without TTO, the excess volume of \methodname{} rises to 201\%, compared
with 54\% on DT4D, and that of SuperFlex to 116\%, compared with 31\%.
Under this domain shift, the feed-forward prediction therefore
over-covers the object, and TTO becomes important: it reduces the excess
volume of \methodname{} to 65\%, below the 70\% of SuperFlex+TTO, while
keeping a TSR of 0.84. TTO also costs more thin-part coverage than on
DT4D, where TSR drops by only one point.

\paragraph{Overall accuracy.}
After TTO, \methodname{} achieves the lowest Hausdorff distance of all
methods (5.78, in units of $10^{-2}D$), including the convex
decompositions, and a directed distance $\mathrm{HD}_{\mathrm{GT}\to\mathrm{p}}$
of 3.96, on par with Thul et al.\ (3.98) and CoACD (3.99). It also
improves over SuperFlex+TTO in IoU (0.75 vs.\ 0.72), F-score (0.74 vs.\
0.68), Chamfer distance (0.86 vs.\ 1.02), and connectivity (1.99 vs.\
2.41 components).

\paragraph{Comparison with convex decompositions.}
As on DT4D, Thul et al.\ and CoACD cover all thin parts (TSR 1.00), but
with 96--141\% excess volume, 59 to 101 times more storage per frame, and
construction times of 20 to 93\,s per frame, compared with 74\,ms for
\methodname{}+TTO.

\paragraph{Scope.}
We evaluate geometry only. Planning on Objaverse would require placing
each character in the rooms of Sec.~\ref{sm:planning}, which we leave to
future work.

\begin{table}[h]
\centering
\small
\caption{Valid Objaverse clips per method. Each clip has 64 frames.}
\label{tab:sm-objaverse-clips}
\begin{tabular}{@{}lrr@{}}
\toprule
Method & Clips & Frames \\
\midrule
\methodname, \methodname+TTO & 99 & 6,336 \\
\twoline{\methodname{} (SuperDec),}{\methodname{} (SuperDec)+TTO} & 99 & 6,336 \\
Thul et al.\ ($K=16$) & 87 & 5,568 \\
Thul et al.\ ($K=32$) & 86 & 5,504 \\
Skinning Capsules & 98 & 6,272 \\
EMS ($K=32$) & 94 & 6,016 \\
Marching Primitives & 90 & 5,760 \\
All other methods & 100 & 6,400 \\
\bottomrule
\end{tabular}
\end{table}

\begin{table}[t]
\centering
\caption{Geometric reconstruction on Objaverse. Top: primitive-based methods;
bottom: TTO variants, convex decompositions, and Skinning Capsules.
\textsc{SuperDyn}$_{\text{SuperDec}}$ applies our method to SuperDec,
+Kabsch denotes per-primitive Kabsch propagation, and numbers in
parentheses give $K$. All SuperFlex variants use fine-tuned weights.
HD is in units of $10^{-2}D$; size (amortized over the clip) and time are
per frame. Unavailable runs are marked --. Bold, underline, and italics mark the
first, second, and third place within each part.}
\label{tab:objaverse}
\sisetup{round-mode=places,round-precision=2,round-pad=false,
         group-digits=none, mode=text,
         reset-text-series=false, reset-text-shape=false}
\footnotesize
\setlength{\tabcolsep}{3pt}
\renewcommand{\arraystretch}{1.02}
\resizebox{\columnwidth}{!}{\begin{tabular}{@{}l|*{7}{r}@{}}
\toprule
Method & TSR$_{0.01}$ & Excess & HD & HD$_{\mathrm{GT}\to\mathrm{p}}$ & Parts & Size & Time \\
 & $\uparrow$ & (\%)\,$\downarrow$ & $\downarrow$ & $\downarrow$ & & (KiB)\,$\downarrow$ & (ms)\,$\downarrow$ \\
\midrule
\multicolumn{8}{@{}l}{\textit{Primitive-based}} \\
\midrule
\methodname & \textbf{\num{0.9076050379184715}} & \num{201.47092015710237} & \textbf{\num{6.807173269939433}} & \textbf{\num{5.9321374581030915}} & \num{31.404040404040405} & \num{1.5313} & \textbf{\num{0.7833683752171523}} \\
\textsc{SuperDyn}$_{\text{SuperDec}}$ & \underline{\num{0.8892264373728488}} & \num{236.02531914707873} & \underline{\num{7.1549378103548165}} & \underline{\num{6.097922881595439}} & \num{31.838383838383837} & \num{1.5156} & \underline{\num{1.0292762207780535}} \\
\midrule
SuperFlex & \num{0.4305965179452256} & \num{116.14547878249213} & \num{10.63513677796171} & \num{10.621194569397339} & \num{8.10984375} & \num{1.7656} & \num{789.1651615232936} \\
SuperFlex+Kabsch & \num{0.4228291771700288} & \num{117.04973211191025} & \num{10.675875011387918} & \num{10.606737201881163} & \num{8.16} & \underline{\num{0.9504}} & \num{12.89411158060829} \\
SuperDec & \num{0.33091456564817173} & \textit{\num{106.84689458879303}} & \textit{\num{10.271079627890545}} & \textit{\num{9.743155796593737}} & \num{12.7046875} & \num{1.7656} & \num{6.855403905137791} \\
SQ & \textit{\num{0.5472657435276675}} & \num{513.0855681724917} & \num{14.131185241548277} & \num{10.139146190174554} & \num{10.0} & \num{2.2070} & \textit{\num{3.988930921186693}} \\
CSA & \num{0.22944375716293738} & \underline{\num{69.44235822878925}} & \num{12.415649874840392} & \num{12.279683300857117} & \num{10.065} & \num{2.4847} & \num{8.979559955041623} \\
EMS (8) & -- & -- & -- & -- & -- & -- & -- \\
EMS (32) & \num{0.44636176472237704} & \num{361.15717700983924} & \num{15.378107802883475} & \num{11.113849406894019} & \num{6.065824468085107} & \textbf{\num{0.5983}} & \num{175.87676140319672} \\
Marching Prim. & \num{0.26678706880969727} & \textbf{\num{10.677187853469382}} & \num{16.03849010223415} & \num{16.03761850894893} & \num{12.928125} & \textit{\num{1.2751}} & 107.3e3 \\
\midrule
\multicolumn{8}{@{}l}{\textit{With TTO, convex, and rig-based}} \\
\midrule
\textsc{SuperDyn}+TTO & \num{0.8369003353820297} & \textbf{\num{65.30968283760281}} & \textbf{\num{5.784186099749888}} & \textbf{\num{3.9644000738186977}} & \num{31.404040404040405} & \textit{\num{1.5313}} & \textit{\num{73.92174897258738}} \\
\textsc{SuperDyn}$_{\text{SuperDec}}$+TTO & \num{0.8192} & \num{92.63} & \underline{\num{5.85}} & \num{4.40} & \num{31.84} & \underline{\num{1.5156}} & \underline{\num{62.49}} \\
SuperFlex+TTO & \num{0.5362892937453028} & \underline{\num{69.81479016133693}} & \num{9.50162938844687} & \num{9.19455243803364} & \num{8.10984375} & \num{1.7656} & 3.73e3 \\
SuperFlex+Kabsch+TTO & \num{0.5038655129810659} & \textit{\num{69.89101252626216}} & \num{10.302420951985075} & \num{9.519035434722174} & \num{8.16} & \textbf{\num{0.9504}} & \textbf{\num{59.47738469199977}} \\
\midrule
Thul et al.\ (8) & -- & -- & -- & -- & -- & -- & -- \\
Thul et al.\ (16) & \underline{\num{0.9993458962619415}} & \num{123.46077890239202} & \num{7.755787170893381} & \num{4.816824668885902} & \num{15.998204022988507} & \num{90.4132} & 20.3e3 \\
Thul et al.\ (32) & \textit{\num{0.9969048895809144}} & \num{96.28369343379649} & \num{7.330493776333392} & \underline{\num{3.9797795481147777}} & \num{31.98655523255814} & \num{124.6508} & 41.1e3 \\
V-HACD (8) & -- & -- & -- & -- & -- & -- & -- \\
V-HACD (16) & \num{0.8798059187894286} & \num{99.94460507904218} & \textit{\num{6.620272794902082}} & \num{5.198041310174093} & \num{16.0} & \num{22.3931} & 1.17e3 \\
V-HACD (32) & \num{0.8641096236723957} & \num{79.72571278240584} & \num{6.926122155871567} & \num{4.320635427595087} & \num{32.0} & \num{34.0298} & \num{973.3479134825757} \\
CoACD (8) & -- & -- & -- & -- & -- & -- & -- \\
CoACD (16) & \textbf{\num{0.9995979633668647}} & \num{141.1572765917728} & \num{6.647703124168626} & \textit{\num{3.985592612935766}} & \num{15.95953125} & \num{154.2365} & 93.2e3 \\
\midrule
Skinning Capsules & \num{0.6054969128197216} & \num{225.7886488735389} & \num{12.356} & \num{9.054} & \num{30.96} & \num{2.9329} & 18.9e3 \\
\bottomrule
\end{tabular}}
\end{table}

\section{Analysis of Thin-Structure Recall}
\label{sm:tsr-radius}

\begin{figure*}[t]
\centering
\includegraphics[width=\textwidth]{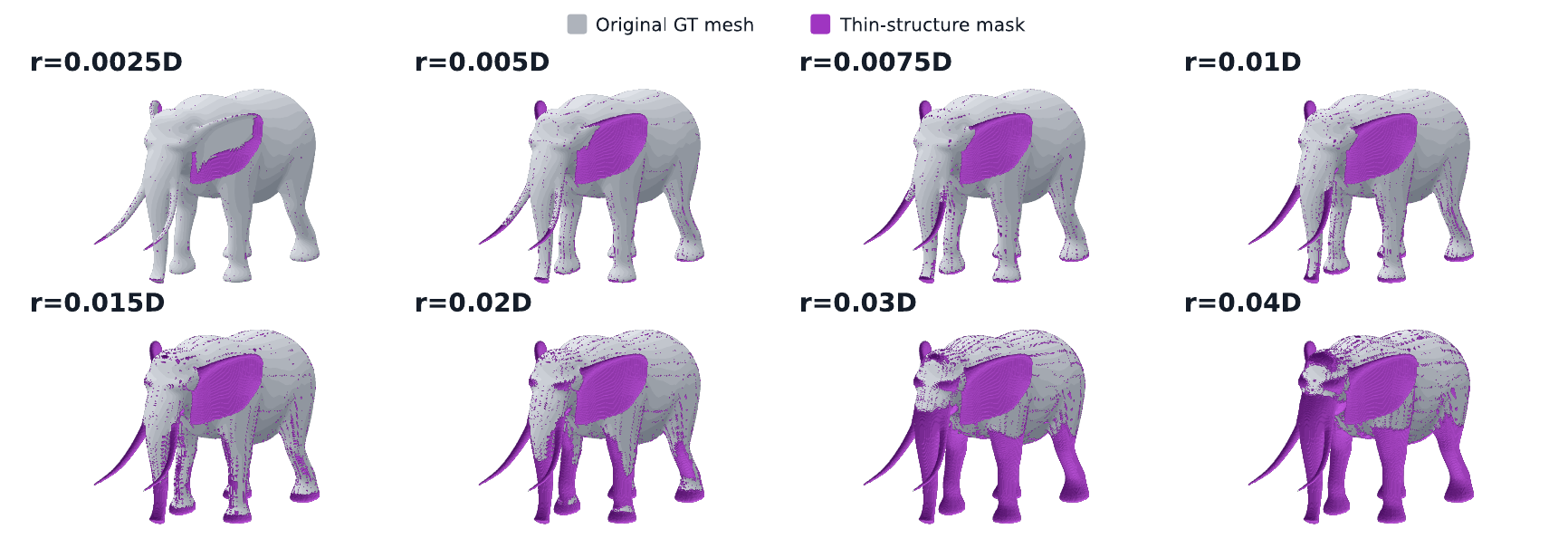}
\captionsetup{hypcap=false}
\captionof{figure}{\textbf{Thin-structure masks at different radii}. The elephant is shown from a fixed oblique viewpoint, with the ground-truth mesh in gray and the visible portions of each thin-structure mask in purple. Each mask is obtained by subtracting the spherical morphological opening at radius \(r\) from the voxelized ground-truth solid. \(D\) denotes the ground-truth bounding-box diagonal. Larger radii capture more of the trunk and legs.}
\label{fig:trs-radii}
\captionsetup{hypcap=true}
\end{figure*}

TSR depends on the radius $r=\rho D$ of the morphological opening
(Eq.~\eqref{eq:tsr}), which sets how thin a part must be to enter
the thin-structure mask $T_r$. Fig.~\ref{fig:trs-radii} shows $T_r$ on an
elephant for $\rho$ between $0.0025$ and $0.04$. At the smallest radii, the
mask mainly contains thin sheets, such as the ears, and the tips of the
tusks. At $\rho=0.01$, it covers the tusks and the tail, and for
$\rho\geq0.03$ it extends over the legs, the trunk, and large parts of the
body, which are no longer thin. Very small radii also approach the voxel
resolution: with 256 voxels along the longest side of the bounding box,
$r=0.005D$ spans only one to two voxels, which makes the mask sensitive to
discretization. We therefore use $\rho=0.01$ in the main paper, which
isolates the parts that primitive-based proxies tend to miss while staying
well above the voxel size.

\begin{figure}
\centering
\includegraphics[width=0.5\textwidth]{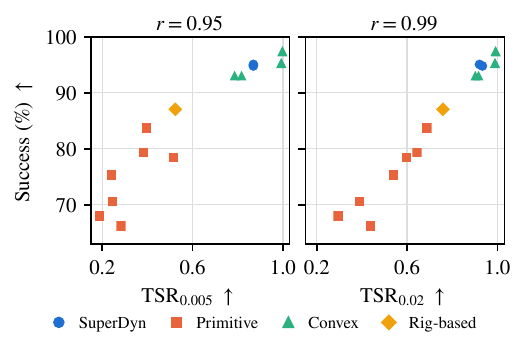}
\captionsetup{hypcap=false}
\captionof{figure}{Planning success vs.\ TSR at other radii on DT4D. Planning success
rate plotted against TSR$_{0.005}$ (left) and TSR$_{0.02}$ (right), where
TSR$_{\rho}$ uses a ball of radius $r=\rho D$ (Sec.~\ref{sec:tsr}). Each point is one
method from Table~\ref{tab:tsr_radius}. The strong correlation holds across radii
(Pearson $r = 0.95$ and $0.99$, respectively), consistent with Fig.~\ref{fig:tsr-iou}, which uses $r = 0.01D$}
\label{fig:corr-radii}
\captionsetup{hypcap=true}
\end{figure}

\begin{figure}[t]
  \centering
  \includegraphics[width=0.8\linewidth]{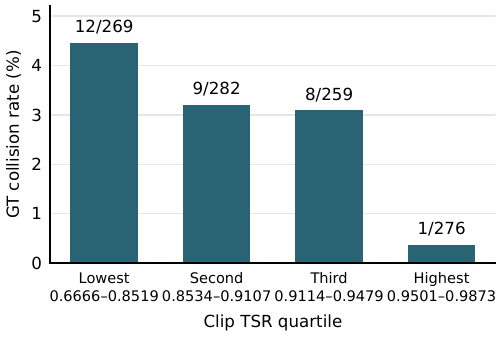}
  \caption{GT collisions among returned paths of \methodname{}, grouped by
    clip TSR quartile, in rooms with a single deforming obstacle. Labels
    give collisions over returned paths.}
  \label{fig:supp_tsr_within}
\end{figure}

The ranking of methods is consistent across radii.
Table~\ref{tab:tsr_radius} reports TSR on DT4D for
$\rho\in\{0.005, 0.01, 0.02\}$. The first three places within each group
are the same at every radius. \methodname roughly doubles the TSR of
SuperFlex at $\rho=0.005$ ($0.87$ vs.\ $0.38$) and remains clearly ahead at
$\rho=0.02$ ($0.93$ vs.\ $0.58$). The gap narrows at larger radii because
the mask then includes thicker parts, which the baselines also cover.
CoACD and Thul et al. stay close to $0.99$ at all radii.
Fig.~\ref{fig:corr-radii} plots planning success against TSR at the two
additional radii. The correlation remains strong, with Pearson $r=0.95$ for
$\rho=0.005$ and $r=0.99$ for $\rho=0.02$, compared with $r=0.97$ for
$\rho=0.01$ on the same set of methods. The value $r=0.96$ in Fig.~\ref{fig:tsr-iou} also includes the ablation variants. TSR thus predicts
planning success at every radius we tested.
\paragraph{TSR within methods.}
The correlation in Fig.~\ref{fig:tsr-iou} compares method averages.
We also relate TSR to collisions within each method. We analyze the
24{,}095 returned paths in rooms~A and~C, which contain a single deforming
obstacle, so that each path is associated with one clip; we exclude runs
without a path, since they reflect over-coverage rather than missed parts.
We fit a logistic model of GT collision on clip-level TSR, controlling for
method, room, IoU, and excess volume, and obtain confidence intervals by
resampling clips.
Across all methods, a 10-point drop in TSR multiplies the odds of a
collision by 1.26 (95\% CI 1.11--1.42; 1.04--1.51 when resampling
characters), so TSR carries information beyond IoU and excess volume.
Within \methodname{}, collisions concentrate in clips with low TSR
(Fig.~\ref{fig:supp_tsr_within}): 4.5\% of returned paths collide in the
lowest TSR quartile, compared with 0.4\% in the highest (odds ratio 1.85
per 10-point drop, CI 1.02--3.33). For \methodname{}+TTO, the trend has the same
direction but is not significant (1.46, CI 0.70--3.01).
Collisions are rare for both variants (30 and 37 paths), and adding TSR
does not significantly improve the prediction of collisions on held-out
clips. These results show an association within methods rather than a per-clip predictor.

\begin{table}[t]
\centering
\caption{Thin-structure recall at different radii on DT4D. TSR$_{\rho}$ uses
a ball of radius $r=\rho D$ (Sec.~\ref{sec:tsr}), so a larger $\rho$ includes thicker
parts in $T_r$. Top: primitive-based methods; bottom: TTO variants, convex
decompositions, and Skinning Capsules. Bold, underline, and italics mark
the first, second, and third place within each part.}
\label{tab:tsr_radius}
\sisetup{round-mode=places,round-precision=2,round-pad=false,
         group-digits=none, mode=text,
         reset-text-series=false, reset-text-shape=false}
\footnotesize
\setlength{\tabcolsep}{6pt}
\renewcommand{\arraystretch}{1.02}
\begin{tabular}{@{}l|*{3}{r}@{}}
\toprule
Method & TSR$_{0.005}$\,$\uparrow$ & TSR$_{0.01}$\,$\uparrow$ & TSR$_{0.02}$\,$\uparrow$ \\
\midrule
\multicolumn{4}{@{}l}{\textit{Primitive-based}} \\
\midrule
\methodname & \textbf{\num{0.8690427768670841}} & \textbf{\num{0.8971453907587048}} & \textbf{\num{0.9334835717393087}} \\
\textsc{SuperDyn}$_{\text{SuperDec}}$ & \underline{\num{0.8403432936591375}} & \underline{\num{0.8714978260645435}} & \underline{\num{0.9163736971848355}} \\
\midrule
SuperFlex & \num{0.38187938659356413} & \num{0.43799667826138117} & \num{0.5758478400661303} \\
SuperFlex+Kabsch & \num{0.37781690495082954} & \num{0.4403029887535068} & \num{0.5859053065205021} \\
SuperDec & \num{0.24432692669995973} & \num{0.2928415608283258} & \num{0.3882905960310503} \\
SQ & \textit{\num{0.5162158360376922}} & \textit{\num{0.5285386476096218}} & \textit{\num{0.5966282259634784}} \\
CSA & \num{0.18754088838947322} & \num{0.22687884528362987} & \num{0.2944338763713633} \\
EMS (8) & \num{0.28123925455319954} & \num{0.3220739369365026} & \num{0.43586087657112116} \\
EMS (32) & \num{0.2831446999663702} & \num{0.3242262750910727} & \num{0.4382872164442787} \\
Marching Prim. & \num{0.2396100641644432} & \num{0.2843438157796177} & \num{0.5408757940551814} \\
\midrule
\multicolumn{4}{@{}l}{\textit{With TTO, convex, and rig-based}} \\
\midrule
\textsc{SuperDyn}+TTO & \num{0.8692994369352759} & \num{0.8882563363692739} & \num{0.9209755732180784} \\
\textsc{SuperDyn}$_{\text{SuperDec}}$+TTO & \num{0.8423739525990043} & \num{0.8676988252874546} & \num{0.9069179025415909} \\
SuperFlex+TTO & \num{0.39493816575643126} & \num{0.5106004416689539} & \num{0.6876982507743498} \\
SuperFlex+Kabsch+TTO & \num{0.3832823449367716} & \num{0.48344477344926345} & \num{0.643516055280501} \\
\midrule
Thul et al.\ (8) & \textit{\num{0.9947264624512349}} & \textit{\num{0.9910808438188254}} & \textit{\num{0.9914160890314085}} \\
Thul et al.\ (16) & \num{0.9936485884481928} & \num{0.9899344606222675} & \num{0.990209335597878} \\
Thul et al.\ (32) & \num{0.9925846054504912} & \num{0.9884877155221252} & \num{0.988854781465241} \\
V-HACD (8) & \num{0.8417615600424386} & \num{0.8795905553111126} & \num{0.9257755073874324} \\
V-HACD (16) & \num{0.8169798832926176} & \num{0.863183918092205} & \num{0.9165370281806104} \\
V-HACD (32) & \num{0.7862922884525844} & \num{0.8409328297291355} & \num{0.9024313236339419} \\
CoACD (8) & \textbf{\num{0.998175794821381}} & \textbf{\num{0.9944737520183312}} & \textbf{\num{0.9932190583243576}} \\
CoACD (16) & \underline{\num{0.9976152291699509}} & \underline{\num{0.9932147328374544}} & \underline{\num{0.991641563371498}} \\
\midrule
Skinning Capsules & \num{0.5235045510534845} & \num{0.6334661239188523} & \num{0.7586843545219725} \\
\bottomrule
\end{tabular}
\end{table}

\section{Temporal Stability}
\label{sec:temporal-stability}

\begin{figure}[t]
\centering
\includegraphics[width=0.45\textwidth]{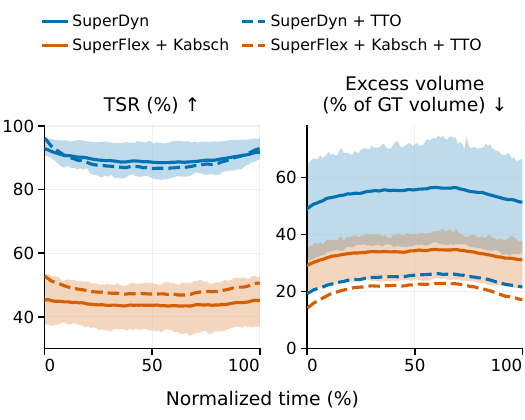}
\caption{\textbf{Temporal stability of rigid propagation.}
TSR$_{0.01}$ (left) and excess volume (right) as a function of the
normalized time since the keyframe, which is the first frame of each
sequence. Each primitive is propagated from the keyframe by per-primitive
rigid alignment (Sec.~\ref{sec:key-frame-to-sequence}). Curves show the mean
over the 200 DT4D test clips, with equal weight per clip; frames are
linearly interpolated onto a common time axis. Shaded regions show the
interquartile range across clips for the feed-forward variants; they are
omitted for the TTO variants for readability. Along the sequence,
\methodname{} retains a large TSR margin over SuperFlex+Kabsch, and its
excess volume increases by a similar amount. TTO raises TSR at the
keyframe, but this gain is specific to the keyframe pose and fades
within the first frames.}
\label{fig:temporal-stability}
\end{figure}

All frames other than the keyframe are obtained by per-primitive rigid
propagation. Fig.~\ref{fig:temporal-stability} shows accuracy as a function of the time since the keyframe. The TSR$_{0.01}$
of \methodname{} remains between approximately 0.88 and 0.95 throughout the
sequence and stays roughly 45 points above SuperFlex+Kabsch at every time
step. Excess volume increases moderately toward the middle of the sequence,
from about 48\% to 55\% of the GT volume, and partly recovers toward the end.
The TSR gain from TTO is confined to the keyframe: after the first few frames,
\methodname{}+TTO falls slightly below the feed-forward model, since the
refined primitives are fitted to the keyframe pose.

\section{Keyframe Selection}
\label{sec:keyframe-selection}

Table~\ref{tab:keyframe-ablation} varies the keyframe from which the sequence is propagated. Without
TTO, the choice has little effect: TSR$_{0.01}$ varies by less than one point,
IoU by less than 0.01, and excess volume by about one point. With TTO, the
middle and random keyframes reduce excess volume from 24.2\% to about 15.5\%
and raise IoU from 0.79 to 0.83, but lower TSR$_{0.01}$ from 0.89 to 0.82.
Refining a keyframe other than the first thus trades thin-structure recall
for a tighter volumetric fit. We keep the first frame as the default, since
thin-structure recall is the property our method targets.

The bottom of Table~\ref{tab:keyframe-ablation} repeats the experiment with global FPS~\cite{qi2017pointnet++} in place of
region-balanced sampling (Table~\ref{tab:sampling-ablation}). Without TTO, this variant is equally
insensitive to the keyframe. With TTO, however, it loses more thin-structure
recall than \methodname{} for every keyframe choice: 9.7 vs.\ 0.9 points for
the first frame, 7.8 vs.\ 7.3 for the middle frame, and 8.5 vs.\ 7.8 for a
random frame. As a result, \methodname{}+TTO reaches a higher TSR$_{0.01}$
than its global-FPS counterpart for all three keyframes. The gap is largest
for the first frame, our default. Random input sampling behaves like global FPS (Table~\ref{tab:sampling-ablation}):
both reach a similar TSR without TTO (0.90), and both lose about ten points
after refinement ($-10.4$ and $-9.7$), whereas region-balanced sampling
loses less than one point.

\begin{table}[t]
\centering
\caption{Keyframe selection on DT4D (200 test clips). The keyframe is the
first retained frame of each clip (our default), its middle frame, or a
frame chosen uniformly at random. All other frames are obtained by
per-primitive rigid propagation from the keyframe. The bottom groups report
the ablated variant with Global FPS sampling. HD is in units of $10^{-2}D$. Unavailable values
are marked --. Bold marks the best value within each group.}
\label{tab:keyframe-ablation}
\sisetup{round-mode=places,round-precision=2,round-pad=false,
         group-digits=none, mode=text,
         reset-text-series=false, reset-text-shape=false}
\footnotesize
\setlength{\tabcolsep}{4pt}
\resizebox{\columnwidth}{!}{\begin{tabular}{@{}l|*{5}{r}@{}}
\toprule
Keyframe & TSR$_{0.01}$ & Excess & IoU & HD & HD$_{\mathrm{GT}\to\mathrm{p}}$ \\
 & $\uparrow$ & (\%)\,$\downarrow$ & $\uparrow$ & $\downarrow$ & $\downarrow$ \\
\midrule
\multicolumn{6}{@{}l}{\textit{\methodname (feed-forward)}} \\
\midrule
First (ours) & \textbf{\num{0.897145}} & \textbf{\num{54.285133}} & \textbf{\num{0.653649}} & \num{7.026381} & \num{6.353405} \\
Middle & \num{0.889078} & \num{55.336178} & \num{0.649462} & \num{6.952018} & \num{6.289280} \\
Random & \num{0.896955} & \num{54.767108} & \num{0.650341} & \textbf{\num{6.873173}} & \textbf{\num{6.201929}} \\
\midrule
\multicolumn{6}{@{}l}{\textit{\methodname+TTO}} \\
\midrule
First (ours) & \textbf{\num{0.888256}} & \num{24.168837} & \num{0.787677} & \num{6.652076} & \num{5.016949} \\
Middle & \num{0.816188} & \num{15.596472} & \num{0.831155} & \textbf{\num{6.292950}} & \num{4.695380} \\
Random & \num{0.819136} & \textbf{\num{15.491715}} & \textbf{\num{0.831888}} & \num{6.362432} & \textbf{\num{4.599366}} \\
\midrule
\multicolumn{6}{@{}l}{\textit{Global FPS (feed-forward)}} \\
\midrule
First & \textbf{\num{0.896493}} & \textbf{\num{53.918075}} & \textbf{\num{0.655896}} & \num{7.075425} & \num{6.330047} \\
Middle & \num{0.885699} & \num{54.949279} & \num{0.651122} & \num{7.107104} & \num{6.452156} \\
Random & \num{0.892797} & \num{54.300134} & \num{0.653234} & \textbf{\num{6.945345}} & \textbf{\num{6.285715}} \\
\midrule
\multicolumn{6}{@{}l}{\textit{Global FPS+TTO}} \\
\midrule
First & \num{0.799471} & \num{15.649912} & \num{0.830237} & \num{6.823855} & \num{4.880783} \\
Middle & \num{0.807590} & \num{15.464150} & \num{0.831908} & \textbf{\num{6.448181}} & \num{4.876166} \\
Random & \textbf{\num{0.808025}} & \textbf{\num{15.229637}} & \textbf{\num{0.833118}} & \num{6.659876} & \textbf{\num{4.744884}} \\
\bottomrule
\end{tabular}}
\end{table}

\begin{table}[t]
  \centering
  \footnotesize
  \setlength{\tabcolsep}{3pt}
  \caption{Ablation of region-balanced sampling on DT4D. We compare random
    sampling, global FPS, and region-balanced sampling before and after 400
    steps of test-time optimization (TTO). HD values are in units of
    $10^{-2}D$. $\Delta$TSR is the change in thin-structure recall after
    TTO, in percentage points. With the first frame as keyframe, region-balanced sampling shows the smallest drop in TSR after refinement.}
  \label{tab:sampling-ablation}
  \begin{tabular}{@{}llcccccc@{}}
    \toprule
    Sampling & TTO
      & \shortstack{IoU\\$\uparrow$}
      & \shortstack{TSR\\$\uparrow$}
      & \shortstack{HD\\$\downarrow$}
      & \shortstack{HD$_{\mathrm{GT}\to\mathrm{p}}$\\$\downarrow$}
      & \shortstack{Excess\\(\%) $\downarrow$}
      & \shortstack{$\Delta$TSR\\(pp)} \\
    \midrule
    Random       &            & 0.65 & 0.90 & 6.99 & 6.41 & 55.12 &          \\
                 & \checkmark & 0.83 & 0.80 & 6.70 & 4.92 & 15.73 & $-10.35$ \\
    \midrule
    Global FPS   &            & 0.66 & 0.90 & 7.08 & 6.33 & 53.92 &          \\
                 & \checkmark & 0.83 & 0.80 & 6.83 & 4.88 & 15.65 & $-9.70$  \\
    \midrule
    Region-bal.  &            & 0.65 & 0.90 & 7.03 & 6.35 & 54.29 &          \\
    (\methodname)& \checkmark & 0.79 & 0.89 & 6.65 & 5.02 & 24.18 & $\mathbf{-0.89}$ \\
    \bottomrule
  \end{tabular}
\end{table}

\section{Statistical Significance of Planning Results}
\label{sec:supp_significance}

We report the statistical significance of the differences in planning success in Table~\ref{tab:planning}. Since all methods are evaluated on the
same scenarios, we use a paired bootstrap~\cite{tibshirani1993introduction}. The three planner seeds of a
room--sequence pair share the same obstacle motion and are therefore strongly
correlated, so we resample the 737 room--sequence pairs rather than the 2{,}211
individual scenarios, keeping all seeds of a pair together. In each of 10{,}000
resamples, we compute the success rate of every method and the difference
between \methodname{}+TTO and each other method on the same resampled pairs. We
report 95\% percentile intervals and two-sided $p$-values, corrected for
multiple comparisons with the Holm procedure~\cite{holm1979simple}. We also repeat the analysis by resampling the 44 test characters instead of
room--sequence pairs.

Table~\ref{tab:significance} shows that \methodname{}+TTO significantly outperforms
SuperFlex+TTO, SuperFlex+Kabsch+TTO, and Skinning Capsules, by 8.0 to 15.7
percentage points. Its advantage over V-HACD is smaller, 2.0 points, but also
significant. We detect no significant difference from Thul et al.\ (32), whose
interval allows a difference in either direction. CoACD (16) achieves a
significantly higher success rate, by 2.4 points, at the cost of
56$\times$ more storage (Table~\ref{tab:planning}). The character-level
bootstrap leads to the same conclusions for all comparisons.

\begin{figure*}[t]
  \centering
  \includegraphics[width=0.75\textwidth]{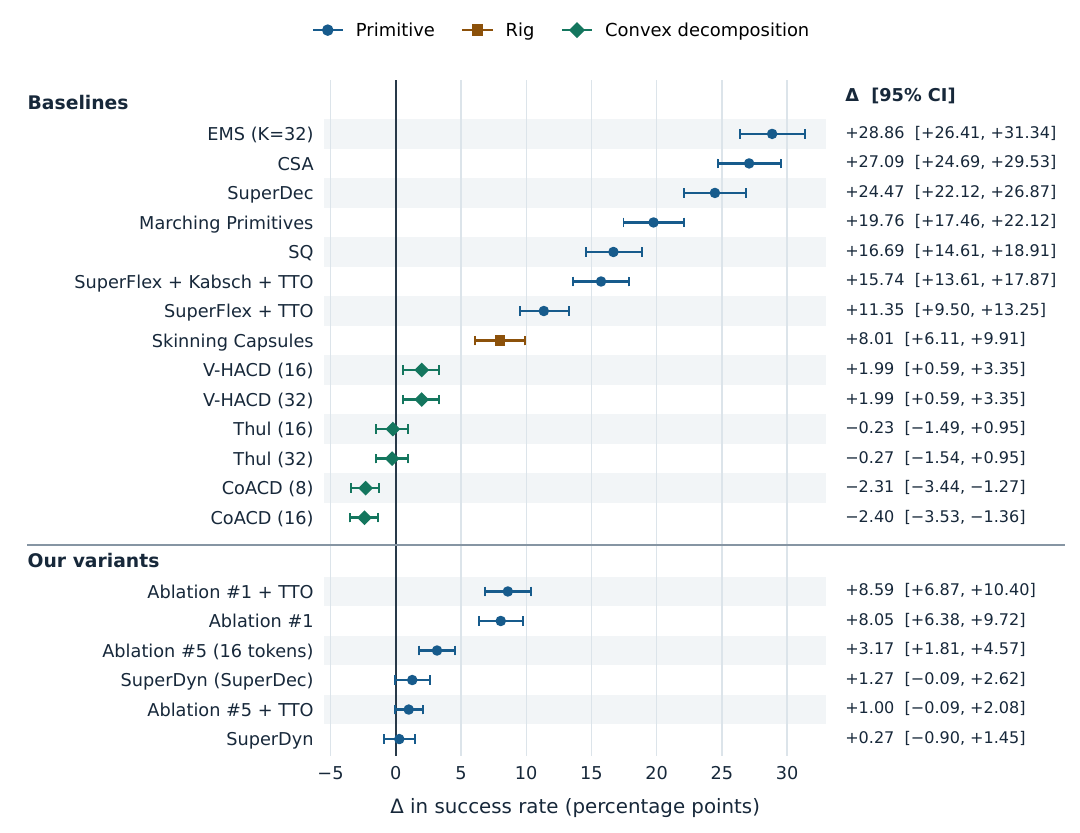}
  \caption{Difference in planning success between \methodname{}+TTO and every
    other method on DT4D, in percentage points. Points show the difference in
    success rate, and bars show pointwise 95\% confidence intervals from
    10{,}000 paired bootstrap resamples of the 737 room--sequence pairs.
    Positive values favor \methodname{}+TTO. Top: baselines, colored by proxy
    type. Bottom: our variants; ablation numbers refer to Table~\ref{tab:ablation-cumulative}. Intervals are not corrected for multiple comparisons;
    Holm-corrected $p$-values for the main baselines are given in
    Table~\ref{tab:significance}.}
  \label{fig:significance_forest}
\end{figure*}

Figure~\ref{fig:significance_forest} extends this comparison to all methods,
including our variants. All primitive-based baselines fall significantly
behind \methodname{}+TTO. Increasing the primitive budget alone (Ablation~\#1)
leaves a significant gap of 8.6 points after TTO, whereas \methodname{} with 16 part
tokens (Ablation~\#5) is not significantly different after TTO, which is consistent
with the gain not stemming from additional capacity. Applied to SuperDec,
\methodname{} is likewise not significantly different from \methodname{}+TTO. 
Finally, feed-forward \methodname{} already achieves 94.80\% planning success
on DT4D, compared with 95.07\% after TTO. The main benefit of TTO is a tighter
geometric fit: excess volume drops from 54.29\% to 24.18\% of the ground-truth
volume, while thin-structure recall remains nearly unchanged (0.90 before
refinement and 0.89 after; Table~\ref{tab:dt4d}).

\begin{table}[t]
  \centering
  \footnotesize
  \setlength{\tabcolsep}{3.5pt}
  \caption{Statistical significance of planning success on DT4D.
    Success rates with 95\% confidence intervals from a paired bootstrap over
    737 room--sequence pairs (10{,}000 resamples), in which all three planner
    seeds of a pair are resampled together. $\Delta$ is the success rate of
    \methodname{}+TTO minus that of the baseline, in percentage points; positive
    values favor our method. $p$-values are two-sided and Holm-corrected over
    the seven comparisons. A bootstrap over the 44 test characters yields the
    same conclusions.}
  \label{tab:significance}
  \begin{tabular}{lccc}
    \toprule
    Method & Success (\%) $\uparrow$ & $\Delta$ (pp) & $p_\text{Holm}$ \\
    \midrule
    \methodname{}+TTO        & 95.07 \ci{93.80}{96.25} & --- & --- \\
    \midrule
    SuperFlex+TTO            & 83.72 \ci{81.73}{85.62} & $+11.35$ \ci{+9.50}{+13.25}  & $<$0.001 \\
    \begin{tabular}[t]{@{}l@{}}SuperFlex+Kabsch\\\quad+TTO\end{tabular}
                             & 79.33 \ci{77.16}{81.50} & $+15.74$ \ci{+13.61}{+17.87} & $<$0.001 \\
    Skinning Capsules        & 87.06 \ci{85.12}{88.92} & $+8.01$ \ci{+6.11}{+9.91}    & $<$0.001 \\
    \midrule
    V-HACD (16)              & 93.08 \ci{91.81}{94.26} & $+1.99$ \ci{+0.59}{+3.35}    & 0.014 \\
    V-HACD (32)              & 93.08 \ci{91.86}{94.26} & $+1.99$ \ci{+0.59}{+3.35}    & 0.014 \\
    Thul et al.\ (32)        & 95.34 \ci{94.26}{96.38} & $-0.27$ \ci{$-$1.54}{+0.95}    & 0.69 \\
    CoACD (16)               & 97.47 \ci{96.56}{98.28} & $-2.40$ \ci{$-$3.53}{$-$1.36}    & $<$0.001 \\
    \bottomrule
  \end{tabular}
\end{table}

\section{Failure Analysis of Planning Runs}
\label{sec:supp_failures}

We classify every planning run as a success, an invalid path, or a missing
path. An invalid path is collision-free for the proxy but collides with the
reference geometry; we further test whether its first contact lies in the
thin-structure mask $T_r$ (Eq.~\eqref{eq:tsr}). A missing path occurs
when the planner returns no solution within 300\,s. For these runs, we also
run a coarse space--time grid search on the proxy, which distinguishes
proxies that block every passage from planner timeouts. One SuperFlex+TTO
run remains unresolved, since its contact with the reference geometry lies
within the 10\,$\mu$m contact tolerance; we count it as a failure.

\begin{figure*}[t]
  \centering
  \includegraphics[width=0.75\linewidth]{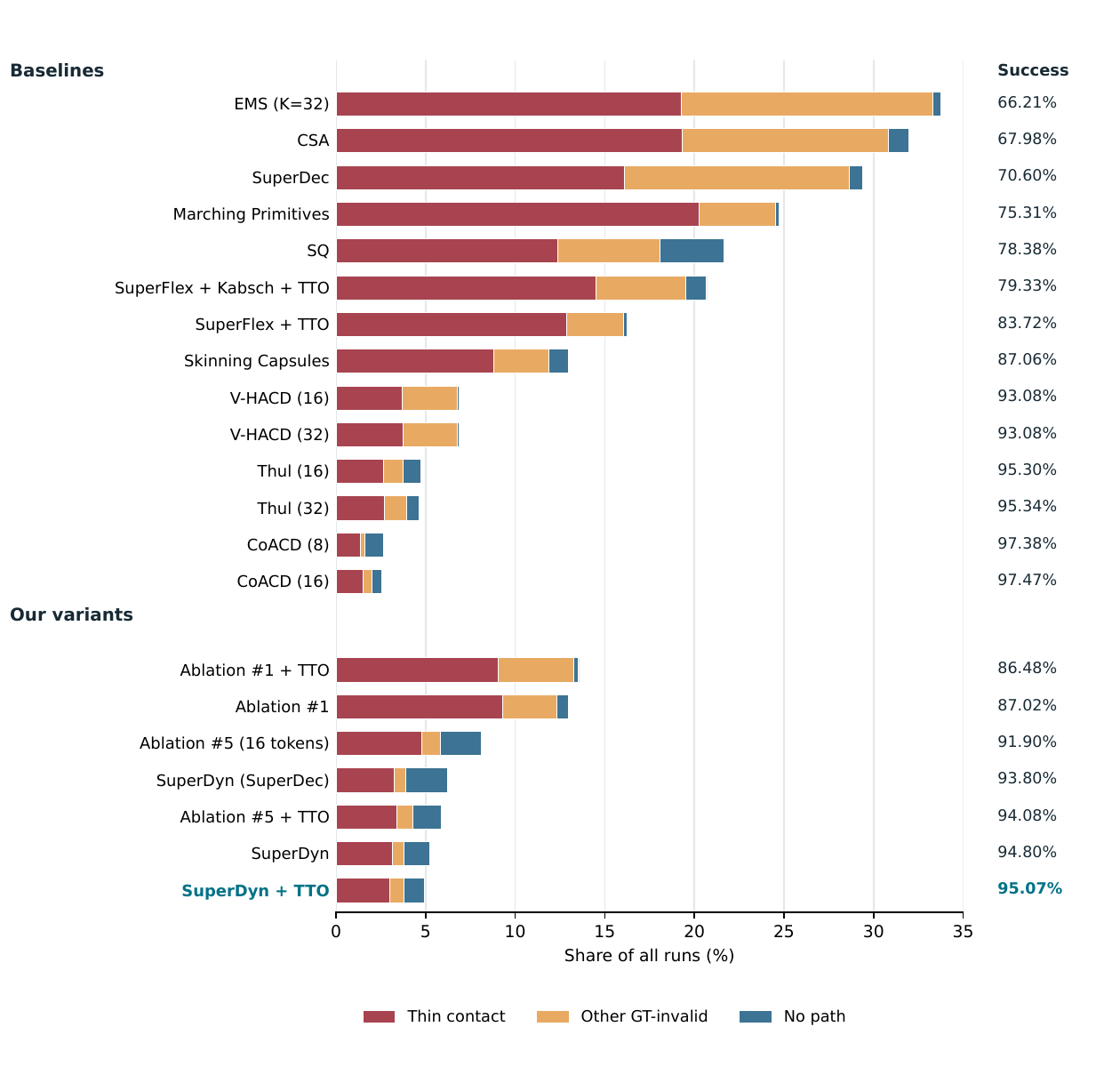}
  \caption{Failure breakdown of the 2{,}211 planning runs per method on DT4D.
    Bars show the share of runs that fail through a first contact in a thin
    region ($T_r$), another collision with the reference geometry, or a
    missing path; the success rate is given on the right. Missed thin parts
    cause most failures of primitive-based methods. One unresolved
    SuperFlex+TTO run (Sec.~\ref{sec:supp_failures}) is counted as a failure
    but not shown.}
  \label{fig:supp_failures}
\end{figure*}

\paragraph{Failure types.}
Figure~\ref{fig:supp_failures} shows that missed thin parts cause most
failures of primitive-based methods.
SuperFlex+TTO returned 355 paths that collided with the ground-truth
geometry, compared with 84 for \methodname{}+TTO. The number of paths
whose first contact occurred in a thin region fell from 284 to 66,
accounting for 218 of the 271 avoided invalid paths (80\%).
Of the 2{,}211 runs, SuperFlex+TTO returned no path in 4 cases
(0.2\%) and \methodname{}+TTO in 25 (1.1\%).
In 11 of these 25 runs, all in room~C, the proxy blocked every passage,
and the grid search also found no path.
In the remaining 14, the grid search found a path, but the planner
timed out.

\paragraph{Rooms.}
Table~\ref{tab:supp_rooms} reports success per room. \methodname{}+TTO
outperforms SuperFlex+TTO by 8.5 to 14.1 percentage points in every room,
including room~C, where most of its missing paths occur. It also outperforms
Skinning Capsules and V-HACD in every room, stays within one point of
Thul et al., and is below CoACD in every room, consistent with
Table~\ref{tab:significance}.

\begin{figure}[t]
  \centering
  \includegraphics[width=\linewidth]{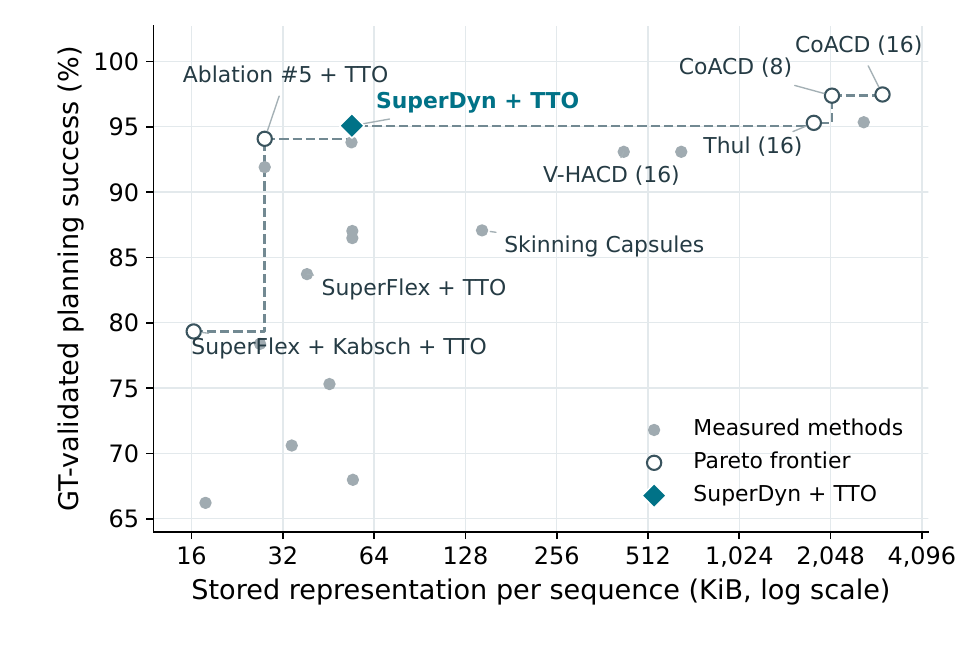}
  \caption{Planning success versus storage per sequence on DT4D (log scale).
    Hollow markers and the dashed line show the Pareto frontier.
    \methodname{}+TTO reaches 95.1\% success with 54\,KiB per sequence; the
    feed-forward \methodname{} has the same storage and lies just below it
    (94.8\%, hidden by the marker).}
  \label{fig:supp_storage}
\end{figure}

\paragraph{Storage.}
Figure~\ref{fig:supp_storage} relates planning success to storage.
\methodname{}+TTO lies on the Pareto frontier: every method with less
storage succeeds less often, and the only methods with higher success,
Thul et al.\ and CoACD, require 33 to 56$\times$ more storage. Safety margins are analyzed separately in Sec.~\ref{sm:safety-margin}.

\begin{table}[t]
  \centering
  \footnotesize
  \setlength{\tabcolsep}{5pt}
  \caption{Planning success (\%) per room on DT4D, with the number of runs per
    room in parentheses. The bottom rows count runs without a returned path.
    \methodname{}+TTO outperforms SuperFlex+TTO in every room.}
  \label{tab:supp_rooms}
  \begin{tabular}{lcccc}
    \toprule
    Method & A (588) & B (525) & C (525) & D (573) \\
    \midrule
    \methodname{}+TTO          & 97.28 & 95.62 & 92.76 & 94.42 \\
    SuperFlex+TTO          & 88.78 & 81.52 & 83.81 & 80.45 \\
    Skinning Capsules      & 90.48 & 85.90 & 87.43 & 84.29 \\
    V-HACD (16)            & 95.41 & 94.48 & 92.00 & 90.40 \\
    Thul et al.\ (32)      & 97.45 & 96.57 & 93.71 & 93.54 \\
    CoACD (16)             & 98.47 & 97.71 & 96.95 & 96.68 \\
    \midrule
    \multicolumn{5}{l}{\textit{No path (runs)}} \\
    \methodname{}+TTO          & 2 & 1 & 15 & 7 \\
    SuperFlex+TTO          & 0 & 0 & 2  & 2 \\
    \bottomrule
  \end{tabular}
\end{table}

\section{Safety Margins for Planning}
\label{sm:safety-margin}

Planning with a safety margin around obstacles is common in robotics. We evaluate how such a margin affects planning with SuperFlex+TTO and \methodname, complementing the dilation experiment in Fig.~\ref{fig:pareto}. During planning, we inflate
the robot radius by a margin of up to 20\,cm. For a spherical robot, this
is equivalent to dilating the proxy by the same amount, without changing
the representation or the collision checker. We still check each returned
path against the original mesh sequence with the true radius of 0.25\,m,
on the same 2,211 scenarios as in the main paper.

\begin{figure*}[t]
\centering
\begin{minipage}[b]{0.48\textwidth}
  \centering
  \includegraphics[width=\linewidth]{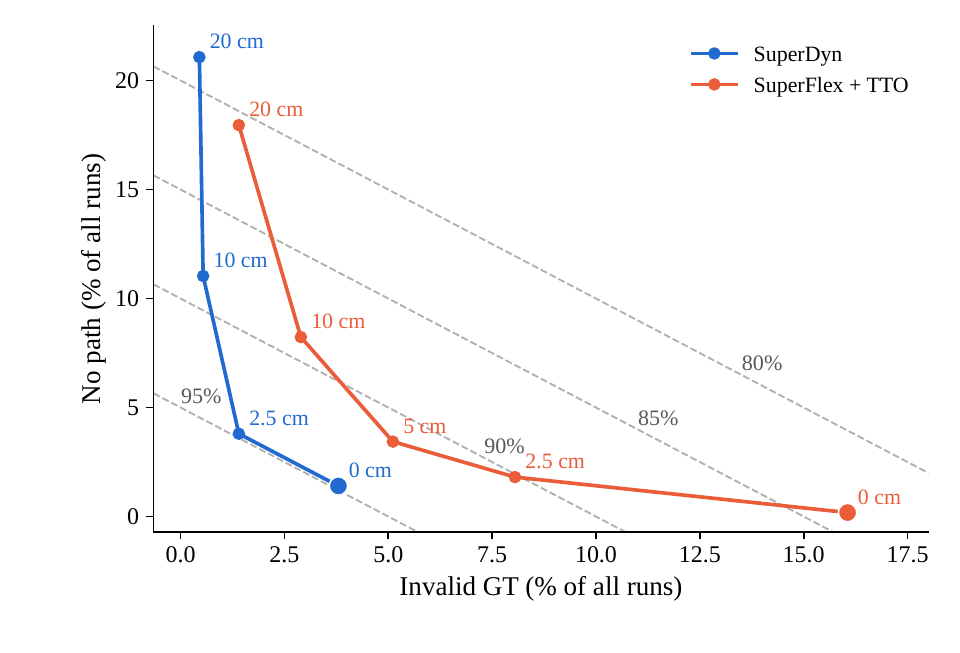}
\end{minipage}\hfill
\begin{minipage}[b]{0.48\textwidth}
  \centering
  \includegraphics[width=\linewidth]{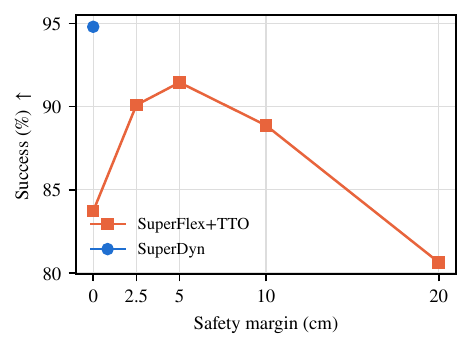}
\end{minipage}
\caption{\textbf{Planning with safety margins.} During planning,
we inflate the robot radius by a margin of 0--20\,cm, which is equivalent
to dilating the proxy, while validity is still checked against the original
mesh sequence with the true radius of 0.25\,m, on the same 2{,}211
scenarios. \methodname{} is the feed-forward variant, and SuperFlex is
refined by per-frame TTO.
\textbf{Left:} invalid GT paths vs.\ no-path failures (\% of all runs);
labels give the margin, large markers show no margin, and dashed lines are
contours of equal success rate. A larger margin trades collisions for
blocked passages. At a similar share of invalid paths, \methodname{} blocks
far fewer passages than SuperFlex+TTO.
\textbf{Right:} planning success vs.\ margin. A tuned margin raises
SuperFlex+TTO from 83.7\% to at most 91.5\% (5\,cm), still below
\methodname{} without any margin (94.8\%). A margin does not improve
\methodname{} (94.8\% at 2.5\,cm) and lowers its success at larger margins.
Invalid GT paths also include collisions caused by obstacle motion between
planning frames, which affect all proxies.}
\label{fig:safety-margin}
\end{figure*}

Fig.~\ref{fig:safety-margin} shows that a margin trades one failure mode
for another. As the margin grows, SuperFlex+TTO returns fewer paths that
collide with the true geometry, but it blocks more passages, so the
planner finds no path in more scenarios. Its success rate peaks at 91.5\%
with a 5\,cm margin and then decreases, remaining below \methodname
without any margin (94.8\%). This peak is also optimistic, since the
margin is selected on the evaluation scenarios themselves. No single margin lets
SuperFlex+TTO reach both the few collisions and the few blocked passages
of \methodname (84 and 31). At a similar number of invalid paths, it
fails to find a path in roughly four times as many scenarios. A margin
does not help \methodname either: its success stays at 94.8\% with
2.5\,cm and decreases at larger margins, as passages are blocked. Even
with a 20\,cm margin, the first recorded contact of 16 invalid
SuperFlex+TTO paths lies on the geometry of the planning frame itself,
compared with one for \methodname. The remaining invalid paths of both
methods come from obstacle motion between planning frames. A missing
thin part is therefore not repaired by a uniform margin, which enlarges
the parts that are already covered rather than the parts that are
missing.

\section{Robustness to Correspondence Errors}
\label{sec:supp_correspondence}

Rigid propagation (Sec.~\ref{sec:key-frame-to-sequence}) relies on point
correspondences, which are exact in our evaluation. To estimate the
effect of imperfect correspondences, as produced by point trackers or
scene flow, we corrupt the target-frame positions $x^{(j)}_n$ while
keeping the keyframe clean. We consider three noise models:
\emph{Gaussian jitter} with a per-axis standard deviation of 0.5--5\% of
$D$, independent across frames;
\emph{outliers}, where 5--40\% of the correspondences in each frame are
replaced by random points on the target surface; and
\emph{random-walk drift}, where the error of each point accumulates over
the frames since the keyframe and reaches an RMS of 1--10\% of $D$ at the
end of the clip.
Besides the standard Kabsch~\cite{kabsch1976solution} solution, we evaluate a robust variant that
minimizes the Huber loss with iteratively reweighted least squares
($\delta = 0.01D$, 12 iterations)~\cite{huber1992robust,holland1977robust}, which down-weights
correspondences with large residuals. We measure geometry on 1{,}000
frames from the 200 test clips with three noise seeds, and planning on
all 737 room--sequence pairs with one planner seed and one noise seed.
Finally, we remove the correspondences altogether and estimate the motion
of each primitive with ICP on independently sampled point clouds.

\begin{figure*}[t]
  \centering
  \includegraphics[width=\textwidth]{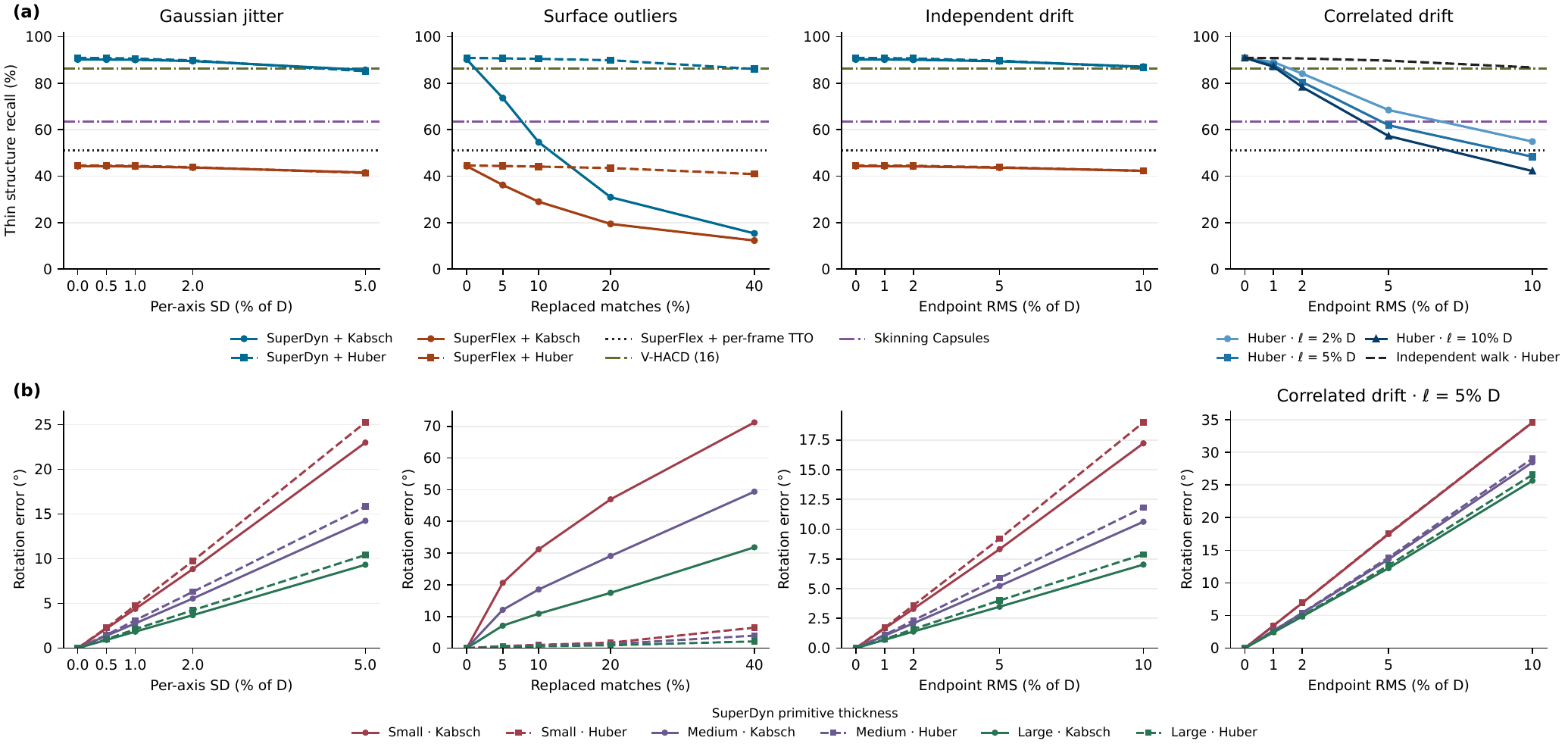}
  \caption{Robustness to correspondence errors on DT4D.
    (a)~Thin-structure recall under Gaussian jitter, outliers, independent
    random-walk drift, and spatially correlated drift with length scale
    $\ell$. The first three panels compare the standard Kabsch solver with
    its Huber variant; the last panel shows the Huber variant, with
    independent drift (dashed) for reference. Horizontal lines show
    reference methods that do not use correspondences.
    (b)~Mean rotation error of \methodname{} primitives, split into thirds by
    primitive thickness; the last panel uses $\ell = 5\%$ of $D$. Thin
    primitives are the most sensitive, and the Huber variant protects them
    from outliers but not from correlated drift. Results are averaged over
    three noise seeds.}
  \label{fig:supp_corr}
\end{figure*}

\paragraph{Geometry.}
Figure~\ref{fig:supp_corr}a shows that \methodname{} is insensitive to jitter
and independent drift: at the largest levels, its TSR remains at 0.85--0.87, compared
with 0.90 without noise. Outliers are different. With the standard Kabsch
solution, 10\% outliers already reduce TSR to 0.55, and 40\% to 0.15. The
Huber variant removes this sensitivity, keeping TSR at 0.86 even with 40\%
outliers, close to V-HACD (16). Without noise, both solvers give nearly
the same TSR (0.90 and 0.91), so the robust variant can replace the standard one at little cost.

\paragraph{Planning.}
Table~\ref{tab:supp_corr} confirms these results in planning. With 40\%
outliers, the success rate of the standard solver drops from 94.8\% to
54.1\%, whereas that of the Huber variant drops by only 1.8 points to
92.5\% and remains 7.5 points above SuperFlex+TTO, which needs no
correspondences. Drift with an RMS of 5\% of $D$ reduces success by
1.1 points, which is not significant.

\paragraph{Thin parts.}
Thin parts are the most sensitive to correspondence errors
(Fig.~\ref{fig:supp_corr}b). Their primitives are elongated, so the
rotation about their axis is poorly constrained, and they contain fewer
points: about 100 per primitive in the thinnest third, compared with
about 210 in the thickest. With 40\% outliers and the standard solver,
their mean rotation error reaches 71$^\circ$, compared with 32$^\circ$ for
the thickest third. The Huber variant reduces these errors to 6.5$^\circ$
and 2.1$^\circ$. Under Gaussian noise, it is slightly less accurate than
the standard solver, as expected for a robust estimator.

\begin{table*}[t]
  \centering
  \footnotesize
  \setlength{\tabcolsep}{3pt}
  \caption{Planning success under correspondence errors on DT4D, over all
    737 room--sequence pairs with one planner seed and one noise seed.
    Intervals are 95\% paired-bootstrap intervals over room--sequence pairs.
    $\Delta_\text{clean}$ compares each configuration with the same solver
    without noise; $\Delta_\text{SF}$ compares it with SuperFlex+TTO on the
    same seed, whose success rate (85.1\%) differs from Table~\ref{tab:planning}, which averages three seeds. For the ICP row, no correspondences
    are used, and $\Delta_\text{clean}$ compares it with the Huber solver on
    exact correspondences.}
  \label{tab:supp_corr}
  \begin{tabular}{llccc}
    \toprule
    Noise & Solver & Success (\%) & $\Delta_\text{clean}$ (pp) & $\Delta_\text{SF}$ (pp) \\
    \midrule
    None          & Kabsch & 94.84 \ci{93.22}{96.34} & ---                          & $+9.77$ \ci{+7.19}{+12.35} \\
    None          & Huber  & 94.30 \ci{92.54}{95.93} & ---                          & $+9.23$ \ci{+6.65}{+11.80} \\
    \midrule
    Outliers 20\% & Kabsch & 72.46 \ci{69.20}{75.58} & $-22.39$ \ci{-25.64}{-19.40} & $-12.62$ \ci{-15.88}{-9.36} \\
    Outliers 20\% & Huber  & 94.44 \ci{92.67}{96.07} & $+0.14$ \ci{-0.68}{+0.95}    & $+9.36$ \ci{+6.92}{+11.94} \\
    Outliers 40\% & Kabsch & 54.14 \ci{50.47}{57.67} & $-40.71$ \ci{-44.37}{-37.04} & $-30.94$ \ci{-34.87}{-27.00} \\
    Outliers 40\% & Huber  & 92.54 \ci{90.64}{94.44} & $-1.76$ \ci{-3.26}{-0.41}    & $+7.46$ \ci{+4.88}{+10.18} \\
    Drift 5\% RMS & Huber  & 93.22 \ci{91.32}{94.98} & $-1.09$ \ci{-2.31}{+0.14}    & $+8.14$ \ci{+5.56}{+10.85} \\
    \begin{tabular}[t]{@{}l@{}}Corr.\ drift 5\%\\\quad($\ell = 5\%$)\end{tabular}
              & Huber  & 86.16 \ci{83.58}{88.60} & $-8.14$ \ci{-10.58}{-5.70}   & $+1.09$ \ci{-2.17}{+4.21} \\
    \midrule
    No corr.\ (ICP) & Huber  & 87.52 \ci{85.07}{89.82} & $-6.78$ \ci{-8.96}{-4.75}    & $+2.44$ \ci{-0.41}{+5.29} \\
    \bottomrule
  \end{tabular}
\end{table*}

\paragraph{Spatially correlated errors.}
The noise above is independent across points, so Kabsch alignment averages
it over the points of each primitive. Point trackers, however, often err
coherently, for example by shifting a whole limb. We therefore also apply
a random-walk field that is correlated in space: 64 control points on the
keyframe carry independent random walks, which we interpolate to all points
with a Gaussian kernel of length scale $\ell \in \{2, 5, 10\}\%$ of $D$ and
rescale to the same endpoint RMS levels as before.
Such errors do not average out (Fig.~\ref{fig:supp_corr}a, right). With
$\ell = 5\%$ of $D$, TSR drops to 0.81 at an RMS of 2\% and to 0.62 at 5\%,
compared with 0.91 and 0.90 for independent drift. Larger length scales,
which move whole primitives together, cause larger drops, and the Huber
variant does not help, since the errors are not outliers. These errors
describe a wrong motion of the part rather than wrong individual matches,
which per-primitive rigid alignment cannot correct. Even at an RMS of
5\%, however, planning success (86.2\%) is not significantly different from
that of SuperFlex+TTO (Table~\ref{tab:supp_corr}), which does not rely on
correspondences.

\paragraph{Without correspondences.}
In every frame, we sample 20{,}000 points uniformly on the surface,
independently of all other frames, so that no point can be matched across
frames. We then estimate a separate rigid transformation for each primitive
by registering its keyframe points to the target cloud. A single rigid ICP
alignment of the whole object first provides a coarse initialization, which
we compose with each primitive's pose from the previous frame. We then refine
each primitive independently with point-to-plane ICP with Huber weights.
This variant keeps a TSR of 0.80, compared with 0.90 with exact
correspondences and 0.51 for SuperFlex+TTO, at the same excess volume as
with exact correspondences (52\%). Its planning success, 87.5\%, is 6.8
points below that with exact correspondences and not significantly
different from that of SuperFlex+TTO (Table~\ref{tab:supp_corr}), while
the decomposition remains temporally consistent. The loss concentrates
on fast and complex motions: 9 of the 23 dance clips lose at least 30
points of TSR, compared with 8 of the other 177 clips. Unlike with
SuperFlex, most of the additional collisions lie outside thin regions
(48 of 78), since ICP errors move whole primitives rather than miss thin
parts. Our unoptimized implementation takes 2.1\,s per frame on a single
CPU core; since the primitives are registered independently, this step
parallelizes directly.

\begin{figure*}[t]
\centering
\includegraphics[width=\linewidth]{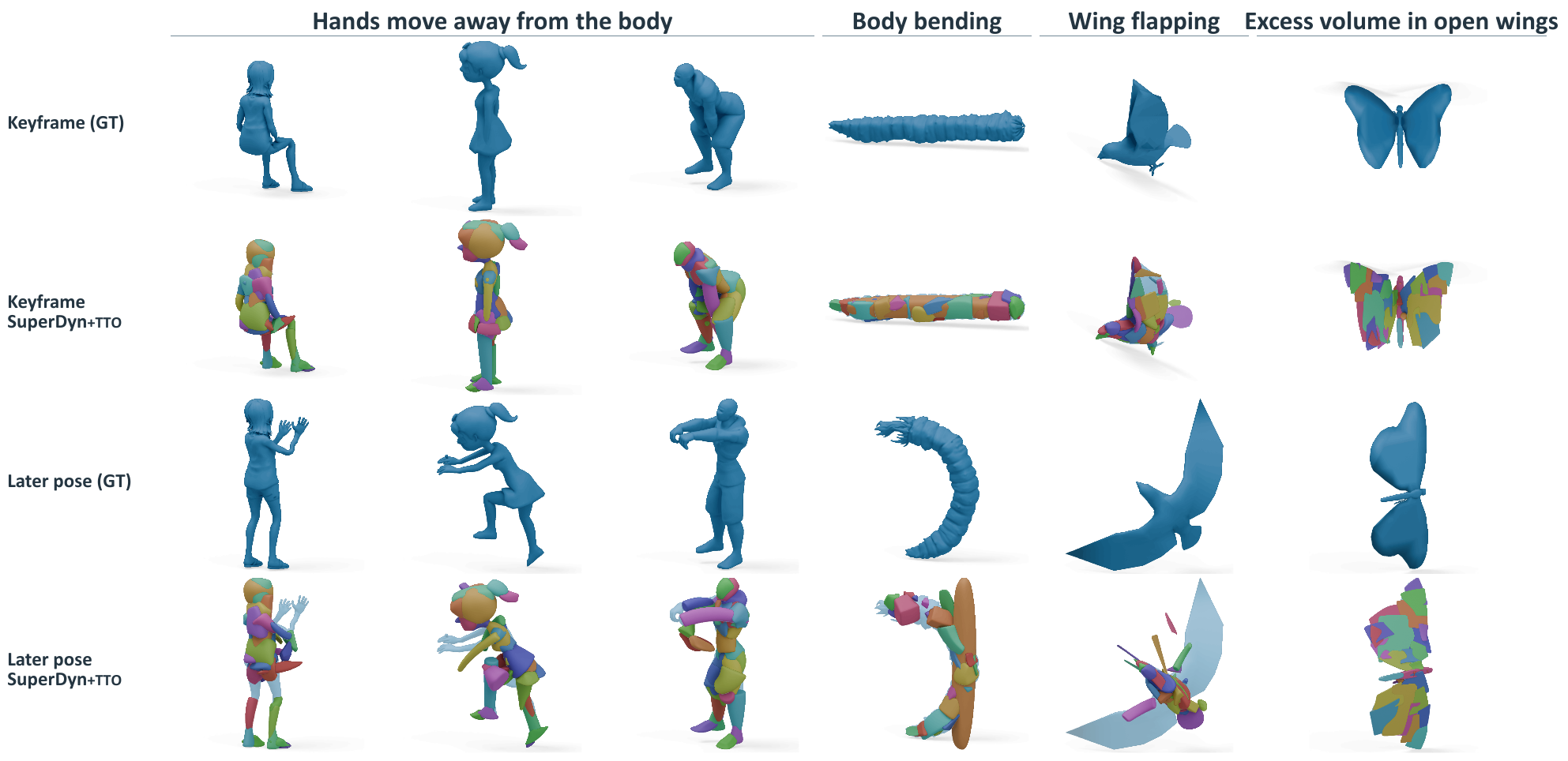}
\caption{\textbf{Failure cases.} Rows show the GT mesh and
\methodname{}+TTO at the keyframe and at a later frame. Semi-transparent
gray regions mark GT geometry that is not covered by any primitive.
\emph{Hands:} the hands touch the body at the keyframe and are covered by
primitives shared with the torso, so they are left uncovered once they move
away. \emph{Body bending:} a large elongated primitive along the body axis
is hidden inside the other primitives at the keyframe and becomes exposed
as the body bends. \emph{Wing flapping:} the wings are folded at the
keyframe, and the reconstruction does not follow them as they spread.
\emph{Open wings:} even when the wings are open at the keyframe, the
primitives are much thicker than the wings, and this excess volume persists
in later frames.}
\label{fig:limitations}
\end{figure*}

\section{Limitations}
\label{sm:limitations}

\paragraph{Point correspondences.}
Rigid propagation requires point correspondences across all frames of a
sequence (Sec.~\ref{sm:correspondences}). In our evaluation, these are exact because the meshes
of each sequence share their vertices. In real captures, they would have to
be estimated, for example by point tracking or scene flow.
Section~\ref{sec:supp_correspondence} shows that propagation tolerates
errors that are independent across points, since Kabsch alignment averages
them over the points of each primitive: independent drift of up to 5\% of
$D$ reduces planning success by about one point, and a Huber-weighted
solver keeps the drop below two points even when 40\% of the
correspondences are wrong. Errors that are correlated across a part, for
example when a tracker shifts a whole limb, do not average out: with a correlation length of 5\%
of $D$, an endpoint error of 5\% of $D$ reduces TSR from 0.91 to 0.62, and
planning success falls to the level of per-frame SuperFlex+TTO
(Sec.~\ref{sec:supp_correspondence}). Thin parts are the most affected,
since their primitives contain fewer points and their rotation about the
long axis is poorly constrained. Without any correspondences,
per-primitive ICP on independently sampled point clouds keeps a TSR of 0.80 and planning success on par with
per-frame SuperFlex+TTO, but loses accuracy on fast motions such as
dancing and adds a registration cost of about 2\,s per frame in our
unoptimized implementation (Sec.~\ref{sec:supp_correspondence}).

\paragraph{Single-frame prediction.}
Although the input sequence contains many frames, the feed-forward model
predicts the decomposition from the keyframe alone, and the remaining
frames are used only for propagation. Information that is visible in other
frames, such as a part that separates from the body later in the sequence,
therefore does not inform the decomposition. Aggregating evidence across
frames is a natural extension.

\paragraph{Keyframe sensitivity.}
Without TTO, average thin-structure recall and excess volume vary little
with the choice of keyframe (Table~\ref{tab:keyframe-ablation}).
After TTO, using the first frame yields higher recall than using a middle
or randomly selected frame (0.89 vs.\ 0.82), but also higher excess volume
(24\% vs.\ about 15.5\% of the ground-truth volume; Sec.~\ref{sec:keyframe-selection}).
On individual clips, the keyframe pose can determine whether a moving
part receives its own primitives. For example, when the hands touch the
torso at the keyframe, the same primitives cover both. Since most of
their assigned points lie on the torso, these primitives follow the torso
as the hands move away, leaving the hands uncovered
(Fig.~\ref{fig:limitations}, left). A similar failure occurs when wings
are folded against the body at the keyframe and spread later in the
sequence (Fig.~\ref{fig:limitations}, wing flapping).

\paragraph{Non-rigid parts.}
Each primitive moves rigidly with its part. When a part bends strongly
during the sequence, primitives that span a large portion of it cannot
follow its shape. On a crawling worm, for example, the decomposition
contains a large elongated primitive along the body axis
(Fig.~\ref{fig:limitations}, body bending). At the keyframe, this primitive
is hidden inside the other primitives, so it does not affect the
reconstruction. As the body bends, it remains straight and protrudes from
the body.

\paragraph{Broad, thin structures.}
Structures that are extremely thin but extend over a large area, such as
butterfly wings, are difficult to represent with a small number of
superquadrics. Even when the wings are open at the keyframe, the primitives
that cover them are much thicker than the wings
(Fig.~\ref{fig:limitations}, open wings). The resulting excess volume
persists throughout the sequence.

\paragraph{Excess volume.}
\methodname{} covers thin parts at the cost of a larger proxy: without TTO,
its excess volume is higher than that of SuperFlex (Table~\ref{tab:dt4d}). In our planning experiments, this rarely blocks a passage, but
all runs in which \methodname{}+TTO blocks every path occur in a single room
(Sec.~\ref{sec:supp_failures}). Environments with many narrow passages
would make this cost more visible. TTO reduces the excess volume while
preserving thin-part coverage, and is preferable in such settings.

\paragraph{Planning benchmark.}
Our planning benchmark uses a spherical robot that moves in the
horizontal plane, in four synthetic rooms with known obstacle motion.
Robots with other shapes, motion in three dimensions, or real scenes with
sensing noise may weigh missed thin parts and excess volume differently.
Our results therefore show that thin-part coverage matters for planning
around deforming obstacles, but they do not establish the size of the
gain in every application.

\end{document}